\pdfoutput=1
\documentclass{article}
\usepackage{ijcai25}

\usepackage{times}
\usepackage{soul}
\usepackage{url}
\usepackage[hidelinks]{hyperref}
\usepackage[utf8]{inputenc}
\usepackage[small]{caption}
\usepackage{graphicx}
\usepackage{amsmath}
\usepackage{amsthm}
\usepackage{booktabs}
\usepackage{algorithm}
\usepackage{algorithmic}
\usepackage[switch]{lineno}

\usepackage{hyperref}
\usepackage{cleveref}
\crefname{figure}{Fig.}{Figs.}
\crefname{table}{Tab.}{Tabs.}
\crefname{equation}{Eq.}{Eqs.}
\crefname{section}{Sec.}{Secs.}
\usepackage{amssymb}
\usepackage[table,xcdraw]{xcolor}

\title{FBQuant: FeedBack Quantization for Large Language Models}

\author{
Yijiang Liu$^1$
\and
Hengyu Fang$^1$\and
Liulu He$^1$\and
Rongyu Zhang$^1$\and
Yichuan Bai$^1$\and\\
Yuan Du$^{1,2}$\And
Li Du$^{1,2,*}$\\
\affiliations
$^1$School of Electronic Science and Engineering, Nanjing University\\
$^2$Interdisciplinary Research Center for Future Intelligent Chips (Chip-X), Nanjing University, Suzhou\\
\emails
\{liuyijiang, hengyufang, heliulu, rongyuzhang, baiyichuan\}@smail.nju.edu.cn,\\
\{yuandu, ldu\}@nju.edu.cn
}

\begin{document}

\maketitle

\begin{abstract}
Deploying Large Language Models (LLMs) on edge devices is increasingly important, as it eliminates reliance on network connections, reduces expensive API calls, and enhances user privacy. However, on-device deployment is challenging due to the limited computational resources of edge devices. In particular, the key bottleneck stems from memory bandwidth constraints related to weight loading. 
Weight-only quantization effectively reduces memory access, yet often induces significant accuracy degradation. 
Recent efforts to incorporate sub-branches have shown promise for mitigating quantization errors, but these methods either lack robust optimization strategies or rely on suboptimal objectives. To address these gaps, we propose FeedBack Quantization (FBQuant), a novel approach inspired by negative feedback mechanisms in automatic control.
FBQuant inherently ensures that the reconstructed weights remain bounded by the quantization process, thereby reducing the risk of overfitting.
To further offset the additional latency introduced by sub-branches, we develop an efficient CUDA kernel that decreases 60\% of extra inference time. 
Comprehensive experiments demonstrate the efficiency and effectiveness of FBQuant across various LLMs. Notably, for 3-bit Llama2-7B, FBQuant improves zero-shot accuracy by 1.2\%.
\end{abstract}

\renewcommand{\thefootnote}{} 
\footnotetext{$*$ denotes the corresponding author.} 
\renewcommand{\thefootnote}{\arabic{footnote}} 

\section{Introduction}
Large Language Models (LLMs)~\cite{touvron2023llama} have demonstrated remarkable capabilities in natural language processing, driving significant advancements in the field. while their extensive parameter counts enable this high-level performance, they also pose substantial deployment challenges related to memory access, storage, and computation. These issues challenges even more critical for personal, localized applications, where privacy concerns often preclude the use of cloud-end LLM APIs, and where most personal devices lack high-performance accelerators (e.g., top-tier GPUs). Furthermore, as AI systems increasingly rely on LLM-based agents~\cite{Talebirad2023MultiAgentCH,Li2023MetaAgentsSI} to handle ever more complex tasks, the computational demands continue to intensify, raising unprecedented barriers to on-device deployment.

\begin{figure}
    \centering
    \includegraphics[width=\linewidth]{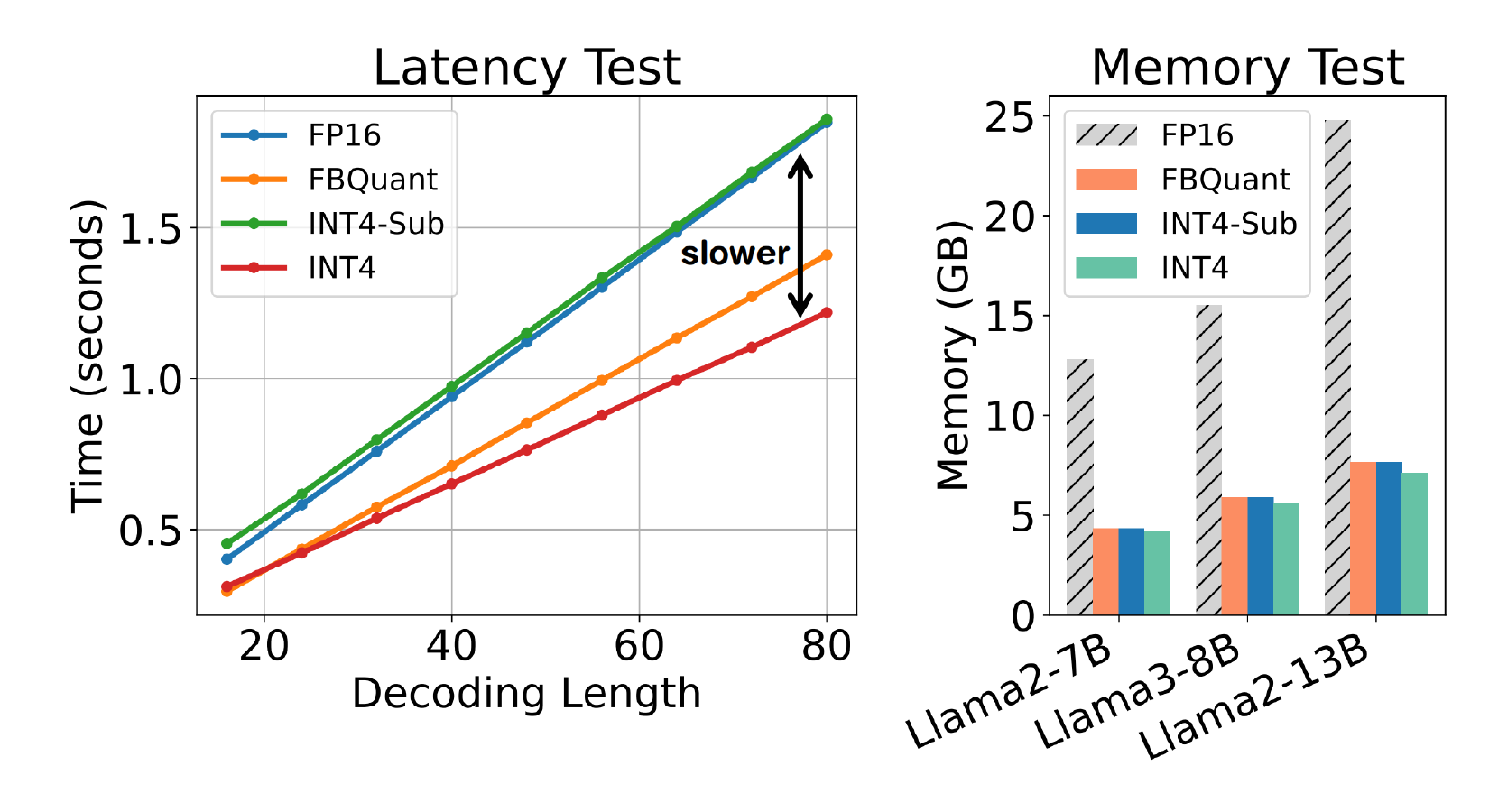}
    \caption{Impact of weight-only quantization on the RTX 3090 GPU. (Left) For Llama2-7B, the INT4 model processes 1,024 tokens for prefilling and 80 new tokens for decoding in only 60\% of the time required by FP16. (Right) After loading to the GPU device, the INT4 model consumes just 25\% of the memory used by FP16.}
    \label{fig:bs1}
\end{figure}

One defining characteristic of the on-device LLMs is the consistently small batch size (in most cases, a batch size of one)~\cite{lin2024awq}. Consequently, inference is predominantly a memory-bandwidth-bound operation constrained by weight loading. Under these conditions, weight-only quantization emerges as a highly effective optimization strategy (as shown in  \cref{fig:bs1}). By storing weights in INT3/4 while retaining inputs and outputs in floating-point, memory requirements and bandwidth overhead are substantially reduced. Moreover, this technique can be broadly applied to various hardware platforms, such as desktops, laptops, and mobile phones, offering a cost-effective solution that balances performance and affordability.

\begin{figure}[t]
    \centering
    \includegraphics[width=1\linewidth]{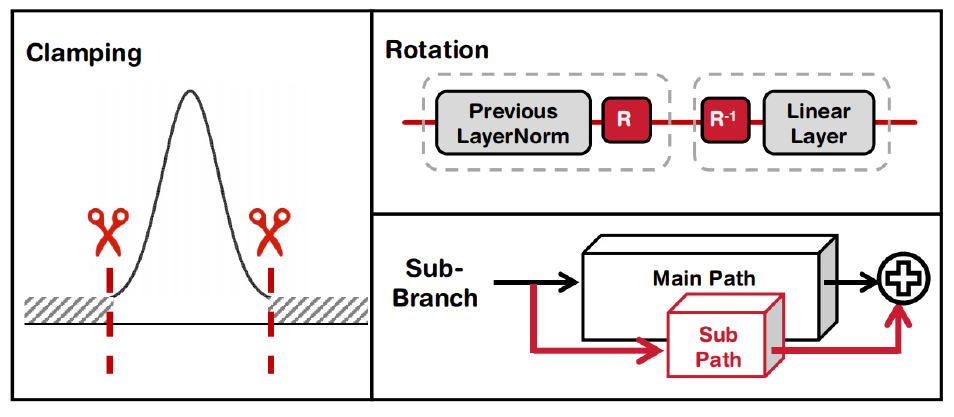}
    \caption{Three categories of optimization methods for weight-only quantization: Clamping, Rotation, and Sub-branching.}
    \label{fig:3wayQuant}
\end{figure}

Existing optimization methods of weight-only quantization for LLMs can be broadly classified into three categories, as illustrated in \cref{fig:3wayQuant}, \textbf{Clamping, Rotation, and Sub-branching}: 
(a) Clamping methods~\cite{lin2024awq,shao2023omniquant} refine the quantization scale by restricting the value range, improving the representation of smaller weights while sacrificing outliers. 
(b) Rotation-based approaches~\cite{liu2024spinquant,lin2024duquant} apply rotation-equivalent transformations to shift quantization challenges from weights (which are quantized) to activations (which remain in high precision).
(c) Sub-branching~\cite{li2023loftq,li2024svdqunat} methods introduce a parallel branch alongside the quantized main path to compensate for quantization errors.
Despite extensive research into clamping and rotation techniques, the quantization-induced accuracy degradation remains a critical concern. Recently, sub-branching methods have gained attention as an orthogonal solution for mitigating quantization errors. The key idea is to split each quantized layer into two parallel paths: a main path that retains the quantized weights and a sub-branch that compensates for quantization losses.
Recent approaches, such as CALDERA~\cite{Saha2024CompressingLL} and EoRA\cite{liu2024eora}, construct sub-branches using the LoRA~\cite{hu2021lora} framework, but these methods can overfit to limited calibration data or rely on ill-posed optimization  objectives (see \cref{sec:motivation}).
Another significant challenge of sub-branching methods is the unexpected increase in inference latency, despite the sub-branches only introducing a small number of operations. 
This delay is primarily driven by memory access bottlenecks, as the sub-branch frequently reads and writes input activations, intermediate results, and layer outputs.

\begin{figure*}[ht]
    \centering
    \includegraphics[width=1\linewidth]{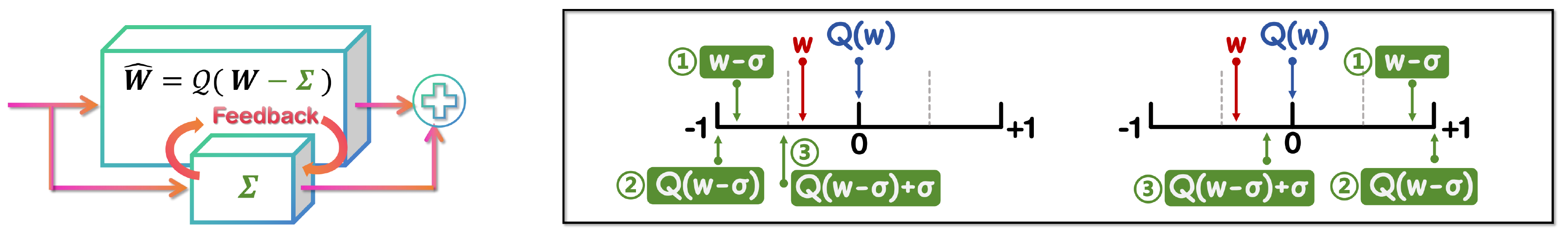}
    \caption{(Left) The main path incorporates feedback signals from the sub-branch to facilitate improved weight quantization, where $\hat{\mathbf{W}}$ represents the quantized weights in the main path, obtained via a quantizer $\mathcal{Q}(\cdot)$, and $\mathbf{\Sigma}$ denotes the weights in the sub-branch. (Right) Direct quantization of the original weights (red) maps them to the nearest quantization bins (blue). In contrast, the FBQuant method (green) applies a multi-step quantization approach, progressively adjusting the weights towards their original values in three stages.}
    \label{fig:framework}
\end{figure*}

In this work, we propose a novel method called \textbf{F}eed\textbf{B}ack \textbf{Quant}ization (\textbf{FBQuant}), tackling both the optimization problem and the latency overhead introduced by sub-branches.
Drawing inspiration from negative feedback mechanisms in automatic control~\cite{franklin2002feedback}, FBQuant feeds the sub-branch weights back into the main path, ensuring that the reconstructed weights remain inherently bounded by the quantization process (see \cref{sec:upper-bound}). With this mechanism, FBQuant is able to fine-tune sub-branches with a limited amount of calibration data, and prevent overfitting.
The pipeline is both straightforward and effective as illustrated in \cref{fig:framework}, and we provide rigorous mathematical proofs in \cref{sec:motivation,sec:method}. 
To address the latency challenge, we develop the CUDA kernel fusion implementation on the sub-branches. By reducing the repeated read and write operations in the constructed layers, this implementation \textcolor{black}{saves 60\% of extra inference time} compared to conventional sub-branches.

Experiments demonstrate that FBQuant outperforms existing methods across various tasks and model families. On 3-bit Llama2-7B, FBQuant improves \textcolor{black}{zero-shot accuracy by 1.2\%}. We further apply FBQuant to instruction-tuned versions of these models, showing consistent gains over other quantization techniques. Wall-clock time evaluations further reveal significant token throughput improvements over traditional implementations, while matching the latency of quantized models without sub-branches.

Our contributions are summarized as follows: 
\begin{itemize}
    \item \textbf{Feedback-Based Sub-Branch Optimization.} We introduce a novel sub-branch quantization pipeline, FBQuant, which feeds sub-branch signals back into the main path to establish more effective reconstruction objectives. This design inherently upper-bounds the reconstructed weights, thereby preventing overfitting to calibration noise.
    \item \textbf{Efficient CUDA Kernel Integration.} We develop a tailored CUDA kernel that fuses the sub-branch operations with the main path, significantly reducing latency by minimizing repeated reads and writes to inputs, intermediates, and outputs. 
    \item \textbf{Superior Performance Across Models.} Extensive experiments demonstrate that FBQuant consistently improves perplexity and zero-shot accuracy over existing quantization approaches across a variety of model families and parameter sizes.
\end{itemize}

\section{Related Works}
\textbf{Quantization Methods} are commonly divided into two categories: Post-Training Quantization (PTQ)~\cite{dong2019hawq,liu2023noisyquant} and Quantization-Aware Training (QAT)~\cite{Liu2023LLMQATDQ}. PTQ offers training-free solutions by reconstructing the quantized model after it has been fully trained. For example, AdaRound~\cite{Nagel2020adaround} optimizes the rounding direction inspired by a Hessian-induced reconstruction objective, while Hawq~\cite{dong2019hawq} leverages Hessian information to identify weights sensitive to quantization. In contrast, QAT incorporates quantization during the training process, mitigating the accuracy loss associated with quantization. In the case of large language models (LLMs), PTQ is generally preferred for its simplicity and lower computational overhead. In this work, we focus on PTQ for LLMs and introduce FBQuant.
\paragraph{Weight-only Quantization} focuses solely on quantizing model weights, reducing memory usage and speeding up memory-bound operations. For instance, \cite{dettmers2022gpt3} introduce the INT8 weight quantization method. QLoRA\cite{dettmers2024qlora} proposes the NF4 data format to better align with LLM weight distributions. GPTQ\cite{frantar2022gptq} uses layer-wise quantization combined with Optimal Brain Compression\cite{frantar2022optimal}, leveraging inverse Hessian information to guide quantization process. AWQ\cite{lin2024awq} identifies salient weights by the corresponding activations and applies scaling techniques to protect them during quantization. OmniQuant\cite{shao2023omniquant} refines quantization via clamping and rotation values learned per layer. However, these methods suffer from significant accuracy degradation under low-bit quantization. In contrast, FBQuant employs an orthogonal sub-branch mechanism that effectively compensates for quantization errors, leading to improved performance.
\paragraph{Weight-Activation Quantization} handles both weights and activations.  SmoothQuant~\cite{Xiao2022SmoothQuantAA} and Outlier Suppression~\cite{Wei2022OutlierSP} achieve W8A8 quantization by balancing the quantization challenges of weights and activations. QServe~\cite{lin2024qserve} employs a W4A8 scheme, offering a trade-off between accuracy and speed. Quarot~\cite{ashkboos2024quarot} and DuQuant~\cite{lin2024duquant} further enhance accuracy by introducing Hadamard rotation on weights and activations. RPTQ\cite{Yuan2023RPTQRP} and LLM-QAT\cite{Liu2023LLMQATDQ} achieve W4A4 quantizaiton.
However, weight-activation quantization generally faces insufficient support on general devices, making weight-only methods more applicable in practice.
\paragraph{Sub-branches Compensation} introduces parallel pathways alongside the quantized layer to offset quantization errors. LoftQ \cite{li2023loftq} directly decompose quantization errors using Singular Value Decomposition and construct low-rank residual paths correspondingly. CALDERA \cite{Saha2024CompressingLL} relies on quantized low-rank sub-branches that are fine-tuned to further reduce quantization errors, while EoRA \cite{liu2024eora} projects compression errors into the eigenspace of input activations, focusing on reconstructing the most impactful components. SVDQuant \cite{li2024svdqunat}, originally developed for diffusion models but adaptable to LLMs, observes that high-rank components capture most of the outliers, leaving the remaining components simpler to quantize. However, these methods lack robust optimization strategies or employ suboptimal reconstruction objectives, often resulting in diminished performance or overfitting to calibration data noise. In contrast, our FBQuant introduces a feedback-driven optimization mechanism that effectively addresses the issue.
\section{Motivation}
This section discusses our insights on the ill-posed optimization of existing sub-branching methods (\cref{sec:moti-1}), and the inference delay induced by the sub-branches (\cref{sec:moti-2}).
\label{sec:motivation}
\subsection{Ill-posed Optimization} 
\label{sec:moti-1}
Sub-branch compensation reconstructs the weights of a layer can be mathematically represented as:
\begin{equation}
    \mathbf{W}'= \mathbf{W}_\mathcal{Q} + \mathbf{\Sigma},
\end{equation}
where $\mathbf{W}$  is the original weight matrix, $\mathbf{W}_\mathcal{Q}$ is its quantized counterpart produced by quantizer $\mathcal{Q}$, and $\mathbf{\Sigma}$ is a low-rank linear projection derived from the sub-branch.
Although this design aims at correcting quantization errors, directly optimizing $\mathbf{\Sigma}$ can lead to overfitting and dependence on calibration data. To see why, let's consider the layer-wise reconstruction objective:
\begin{equation}
\begin{aligned}
\mathbf{\Sigma}^* = \arg\min_{\mathbf{\Sigma}}\mathcal{L}_1,& \quad \text{subject to} \quad \text{rank}(\mathbf{\Sigma}) \leq r,
\end{aligned}
\label{eq:loss-old}
\end{equation}
where
\begin{equation}
\begin{aligned}
    \mathcal{L}_1 &= \|(\mathbf{W} - \mathbf{W}')\mathbf{X}^\top \|_F\\
    &= \|(\mathbf{W} - \mathbf{W}_{\mathcal{Q}} - \mathbf{\Sigma})\mathbf{X}^\top \|_F,
\end{aligned}
\end{equation}
$\|\cdot\|_F$ denotes the Frobenius norm, $\mathbf{X}$ represents layer inputs, and $r$ is the rank constraint for the sub-branch. Letting  $\mathbf{\Delta} = \mathbf{W}-\mathbf{W}_\mathcal{Q}$, assume $\mathbf{\Sigma}^*$ yields a minimal value $\epsilon_1$ for $\mathcal{L}_1$, then 
\begin{equation}
\begin{aligned}
    \epsilon_1 &= \|(\mathbf{\Delta} -\mathbf{\Sigma}^*) \mathbf{X}^\top\|_F\\
    &=\text{tr} \left( (\mathbf{\Delta} -\mathbf{\Sigma}^*) \mathbf{X}^\top \mathbf{X} (\mathbf{\Delta} -\mathbf{\Sigma}^*)^\top \right),
\end{aligned}
\end{equation}
where $\text{tr}(\cdot)$ denotes the trace operation.
In addition, from the Singular Value Decomposition (SVD), $\mathbf{\Sigma}^*$ can be expanded in terms of the top-$r$ singular vectors:
\begin{equation}
    \mathbf{\Sigma}^*=\mathbf{U}_r\mathbf{S}_r\mathbf{V}_r.
\end{equation}

We can subsequently consider an alternative solution $\mathbf{\Sigma}'$:
\begin{equation}
    \mathbf{\Sigma}'=\mathbf{\Sigma}^* + \mathbf{\Sigma}_N,\quad \text{where}\quad \mathbf{\Sigma}_N = \mathbf{U}_r\mathbf{S}_r(\alpha\mathbf{N}_r),
\end{equation}
with $\mathbf{N}_r$ sharing the same dimensionality as $\mathbf{V}_r$, and a scalar $\alpha$. Because typical calibration data are limited, $\mathbf{X}^\top\mathbf{X}$ is positive semidefinite but not strictly full-rank~\cite{huang2024billm}, implying there exists non-zero $\mathbf{N}_r$ orthogonal to $\mathbf{X}^\top\mathbf{X}$, that is,
\begin{equation}
    \mathbf{N}_r\mathbf{X}^\top \mathbf{X} = \mathbf{0}.
\end{equation}

Then,
\begin{equation}
\begin{aligned}
    \mathbf{\Sigma}_N\mathbf{X}^\top \mathbf{X} &=\mathbf{U}_r\mathbf{S}_r(\alpha\mathbf{N}_r\mathbf{X}^\top \mathbf{X})\\
    &= \mathbf{0}.
\end{aligned}
\end{equation}

Hence, the construction loss $\epsilon'$ which is achieved by $\mathbf{\Sigma}'$ is equal to $\epsilon_1$:
\begin{equation}
\begin{aligned}
    \epsilon' &= \text{tr} \left( (\mathbf{\Delta} -\mathbf{\Sigma}') \mathbf{X}^\top \mathbf{X} (\mathbf{\Delta} -\mathbf{\Sigma}')^\top \right)\\
    &= \text{tr} \left( (\mathbf{\Delta} -\mathbf{\Sigma}^*) \mathbf{X}^\top \mathbf{X} (\mathbf{\Delta} -\mathbf{\Sigma}^*)^\top \right) \\
    &\quad \quad  \color{gray}-\text{tr} \left(  (\mathbf{\Delta}-\mathbf{\Sigma}^*)(\mathbf{\Sigma}_N \mathbf{X}^\top \mathbf{X} )^\top \right) \\
    &\quad \quad  \color{gray}-\text{tr} \left( (\mathbf{\Sigma}_N \mathbf{X}^\top \mathbf{X}) \cdot (\mathbf{\Delta} -\mathbf{\Sigma}^*-\mathbf{\Sigma}_N)^\top \right) \\
    &= \epsilon_1,
\end{aligned}
\label{eq:not-unique}
\end{equation}
where gray terms are $\mathbf{0}$ due to orthogonality. 
This equation demonstrates that $\mathbf{\Sigma}'$ is also a valid solution.
However, for any $w\in \mathbf{W}$, which is reconstructed by $\sigma' \in \mathbf{\Sigma}'$, the difference between the original and the reconstructed weights is:
\begin{equation}
\begin{aligned}
    |w - w'| &= \big|w-\big(\mathcal{Q}(w)+\sigma'\big)\big| \\
    &=|w - \mathcal{Q}(w) - \sigma^* - \alpha\sigma_N|.
\end{aligned}
\end{equation}

With determined $\{w,\mathcal{Q},\sigma^*\}$, the unbounded term $\alpha\sigma_N$ may significantly deviate the reconstructed weights, leading to less meaningful values, which highlights the potential for overfitting and performance degradation. 

\subsection{Delay Attribution of Sub-branches}
\label{sec:moti-2}
\begin{figure}[t]
    \centering
    \includegraphics[width=\linewidth]{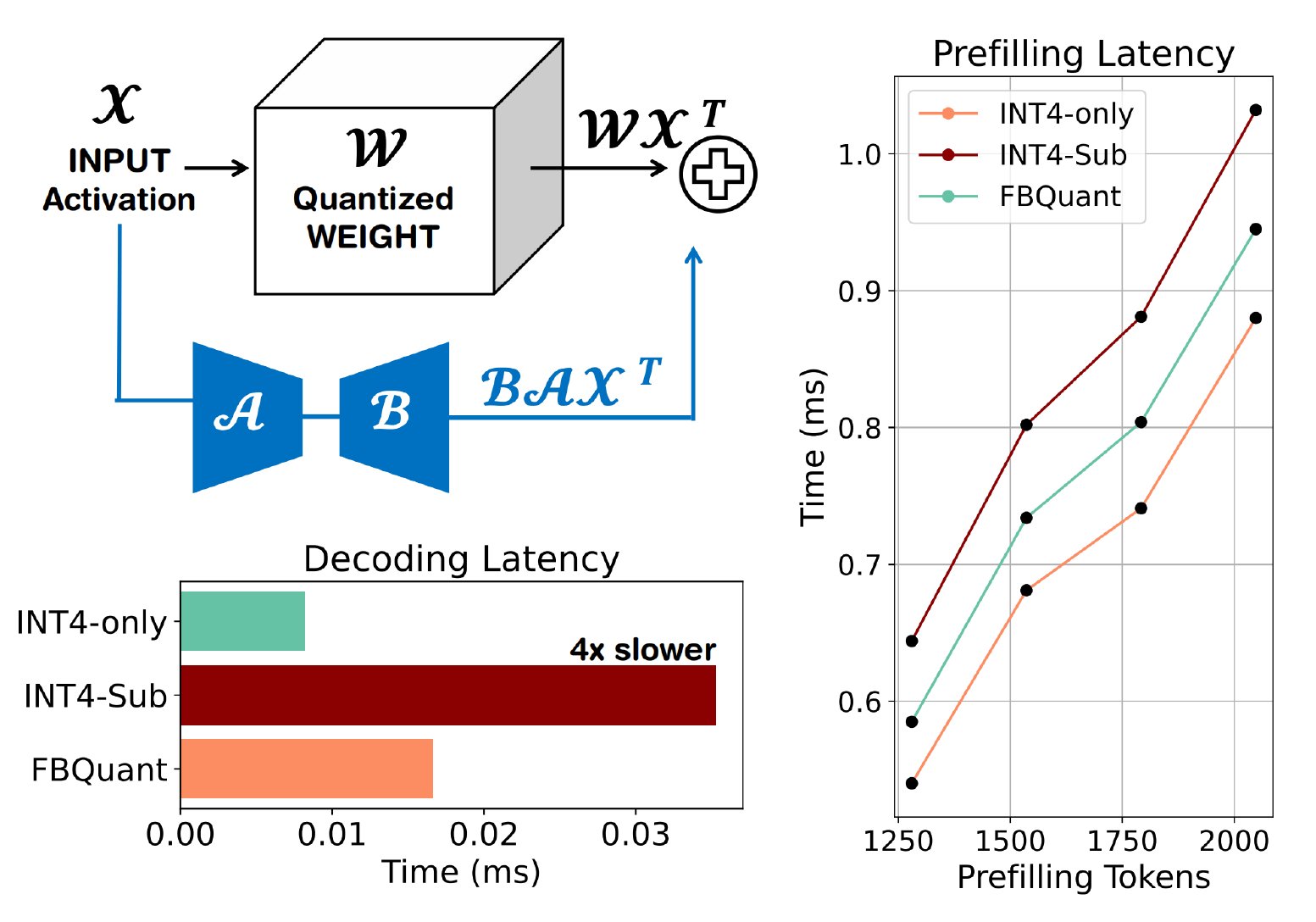}
    \caption{Macs and latency of the linear layer in Llama2-7B. (Up-left) The MACs introduced by the main path $\mathbf{WX}^\top$ and the sub-branch $\mathbf{BAX}^\top$ are $M_0=b\times d\times d$ and $M_1=2\times b\times r\times d$, respectively, where $b$ is the batch size, $r$ is the rank value, and $d$ is the layer dimension.  
    This results in $M_1/M_0=6.25\%$ additional MACs, when $r=128$ and $d=4096$. However, naively implementing this sub-branch significantly increases the latency by 20\% when prefilling (right), and up to four times when decoding (bottom-left). FBQuant significantly mitigates the problem caused by the kernel fusion approach. 
    }
    \label{fig:macs-latency}
\end{figure}

We observe that inference latency can be significantly impacted by the inclusion of sub-branches. Theoretically, the computational overhead introduced by sub-branches constitutes only a small fraction of the total computation in the main path. However, in practice, it can lead to a substantial increase in inference latency. As illustrated in \cref{fig:macs-latency}, consider a linear layer in the Llama2-7B model with a weight matrix $\mathbf{W} \in \mathbb{R}^{d \times d}$, a layer input and output dimension of $d=4096$, and a sub-branch implemented by the LoRA~\cite{hu2021lora} approach with a rank value of 128. The inclusion of the sub-branch results in a 6.25\% increase in Multiply-Accumulate Operations (MACs), but \textcolor{black}{causes the decoding process to be four times slower inside the layer}. This dramatic slowdown is primarily attributable to memory access bottlenecks.
The sub-branch computations involve intensive access to input activations in the down projection (i.e., $\mathbf{A}$), multiple writes to intermediate results before the up projection (i.e., $\mathbf{B}$), and additional writes to the layer outputs.
\begin{figure}[ht]
    \centering
    \includegraphics[width=1\linewidth]{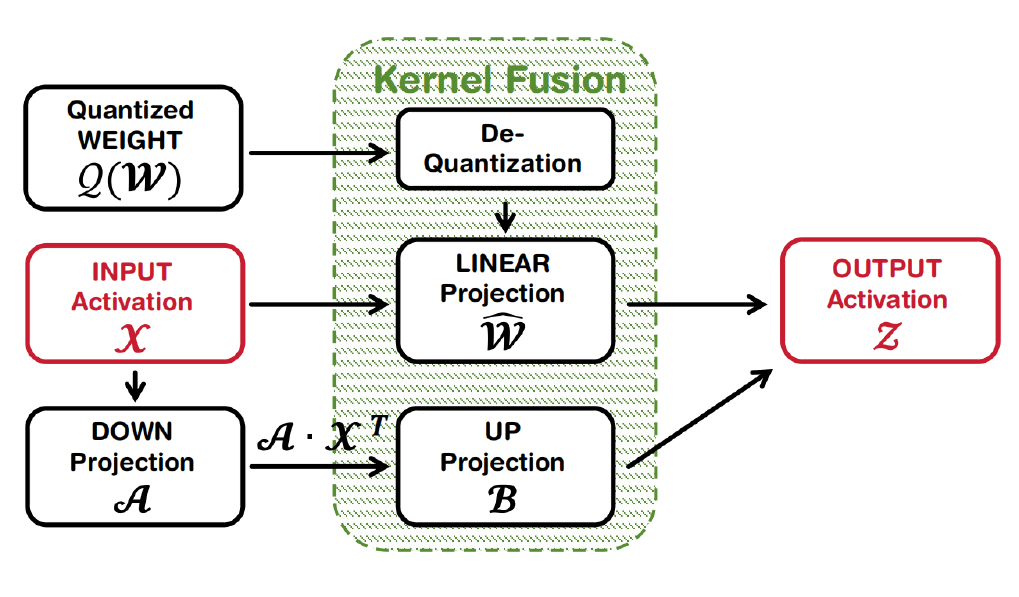}
    \caption{\textbf{Kernel Fusion.} We integrate the de-quantization and the linear projection in the main path, and up-projection in the sub-branch into the same kernel. The reduced number of kernels results in reduced kernel launch time. The integration reduces repeated writes to output activations.}
    \label{fig:kernel-fusion}
\end{figure}

\section{Methodology}
\label{sec:method}
Building on the insights from \cref{sec:motivation}, we propose FBQuant, which includes: (1) feedback integration to upper-bound reconstructed weights and prevent overfitting (see \cref{sec:upper-bound}), and (2) a differentiability strategy ensuring gradient propagation for the sub-branch parameters to process layer-wise reconstruction (see \cref{sec:differentiate}). (3) Furthermore, we introduce an efficient CUDA kernel implementation to mitigate the extra latency induced by the sub-branch (see \cref{sec:kernel}).
\subsection{Feedback and Upper Bound}
\label{sec:upper-bound}
Conventional sub-branch methods overlook on how far the reconstructed weights can deviate from the original. 
To address this limitation, we introduce a novel feedback mechanism, hereafter referred to as \textbf{FBQuant}, which incorporates negative sub-weights into the quantization process. As illustrated in \cref{fig:framework}, the weights of the main path are $\mathcal{Q}(\mathbf{W}-\mathbf{\Sigma})$, and the weights of the sub-branch are $\mathbf{\Sigma}$, then, the feedback-based reconstructed weights $\mathbf{W}_F$ are redefined as:
\begin{equation}
    \mathbf{W}_F = \mathcal{Q}(\mathbf{W}-\mathbf{\Sigma})+\mathbf{\Sigma}.
\end{equation}
Letting $w$ and $\sigma$ be elements of $\mathbf{W}$ and $\mathbf{\Sigma}$ respectively, the difference between the original and reconstructed weights is defined as:
\begin{equation}
\begin{aligned}
    |w - w_F| &= \big|w - \big(\mathcal{Q}(w-\sigma)+\sigma\big)\big| \\
    &=  |(w - \sigma) - \mathcal{Q}(w-\sigma)|.
\end{aligned}
\label{eq:diff}
\end{equation}
When $\mathcal{Q}(\cdot)$ functions as round-to-nearest, the deviation in \cref{eq:diff} is bounded as:
\begin{equation}
    |w - w_F| \leq s/2,
\end{equation}
where $s$ is the quantizer’s scaling factor. Some toy examples are shown in \cref{fig:framework} (left).
Consequently, FBQuant naturally limits reconstructed-weight deviations, which is a critical property that prevents the sub-branch from overfitting calibration data noise.
\subsection{Differentiability for Layer-wise Reconstruction}
\label{sec:differentiate}
We fine-tune sub-branch weights based on layer-wise reconstruction. Before that, we must ensure the differentiability of the reconstruction objectives. Let's first look at the loss function derived by FBQuant:
\begin{equation}
\label{eq:FBQloss}
\begin{aligned}
    \mathcal{L}_F &= \| \mathbf{WX}^\top - \mathbf{W}_F\mathbf{X}^\top \|_F \\
    &= \text{tr}\left( \mathbf{\Delta}_F \mathbf{X}^\top\mathbf{X} \mathbf{\Delta}_F^\top \right),
\end{aligned}
\end{equation}
where
\begin{equation}
\begin{aligned}
    \mathbf{\Delta}_F &= \mathbf{W} - \mathbf{W}_F \\
    &= \mathbf{W} - \mathcal{Q}(\mathbf{W}-\mathbf{\Sigma}) - \mathbf{\Sigma}.
\end{aligned}
\end{equation}

Although Straight-Through Estimator (STE)~\cite{bengio2013estimatingSTE} is commonly used to approximate the derivative of the quantizer as 
\begin{equation}
    \frac{\partial\mathcal{Q}(\mathbf{W})}{\partial \mathbf{W}} \approx \mathbf{I},
\end{equation}
it yields zero gradients for the sub-branch weights $\mathbf{\Sigma}$ in FBQuant, that is
\begin{equation}
\begin{aligned}
    \frac{\partial \mathcal{L}_F}{\partial \mathbf{\Sigma}} &= \frac{\partial \mathcal{L}_F}{\partial \mathbf{\Delta}_F} \cdot \frac{\partial \mathbf{\Delta}_F}{\partial \mathbf{\Sigma}} \\
    &= (2 \cdot \mathbf{\Delta}_F\mathbf{X}^\top\mathbf{X}) \cdot (\mathbf{0} + \mathbf{I} - \mathbf{I}) \\
    &= \mathbf{0}.
\end{aligned}
\end{equation}

To overcome this issue, we detach the feedback signal from the back-propagation graph, while allowing gradients to flow only through the sub-branch. 
This modification yields 
\begin{equation}
    \frac{\partial \mathbf{\Delta}_F}{\partial \mathbf{\Sigma}}=-\mathbf{I}.
\end{equation}

Then, we obtain:
\begin{equation}
\begin{aligned}
    \frac{\partial \mathcal{L}_F}{\partial \mathbf{\Sigma}}  = -2 \cdot \mathbf{\Delta}_F\mathbf{X}^\top\mathbf{X},
\end{aligned}
\end{equation}
which enables optimizing $\mathbf{\Sigma}$ by gradient descent. In practice, we implement $\mathbf{\Sigma}$ using low-rank adapters~\cite{hu2021lora}, such that $\mathbf{\Sigma} = \mathbf{B} \cdot \mathbf{A}$.
The sub-branches are applied to the linear layers in LLMs, such as Query, Key, Value, and Out projections in attention blocks, as well as Down, Gate, and Up projections in feed-forward networks (FFNs). During the optimization process, calibration data is fed into the model and each layer is optimized using gradient descent to minimize the reconstruction loss in \cref{eq:FBQloss}. More details of layer-wise reconstruction by FBQuant are described in \cref{alg:fbquant}.
\begin{algorithm}[t]
\caption{Layer-wise Reconstruction by FBQuant}
\label{alg:fbquant}
\begin{algorithmic}[1]
\REQUIRE 
    $\mathbf{X}_l$: input activations of the $l$-th layer \\
    $\mathbf{W}_l$: original weights of the $l$-th layer \\
    $\mathcal{Q}$: linear quantizer \\
    $\mathbf{\Sigma}_l$, $\mathbf{A}_l$, $\mathbf{B}_l$: $\mathbf{\Sigma}_l \gets \mathbf{B}_l \mathbf{A}_l$, weights forming the low-rank sub-branch \\
    $r$: rank of the sub-branch satisfies $\text{rank}(\mathbf{\Sigma}) \leq r$ \\
\ENSURE
    $\mathbf{A}_l'$, $\mathbf{B}_l'$: optimized low-rank sub-branch weights
\STATE $\mathbf{A}_l \gets \text{instantiate}~\mathbf{A}_l\sim \mathcal{N}(\mathbf{0}, \sigma^2 \mathbf{I})$;
\STATE $\mathbf{B}_l \gets \text{instantiate} ~\mathbf{B}_l ~\text{with} ~\mathbf{0}$;
\FOR{$l = 1 \xrightarrow{} N$} 
    \REPEAT
        \STATE obtain FP16 output: \quad $\mathbf{Z}_l \gets \mathbf{W}_l \mathbf{X}_l^\top$ 
        \STATE obtain quantization output: \\
        \quad \quad $\mathbf{Z}_l' \gets \mathcal{Q}(\mathbf{W}_l - \mathbf{B}_l \mathbf{A}_l) \mathbf{X}_l^\top + \mathbf{B}_l \mathbf{A}_l \mathbf{X}_l^\top$ 
        \STATE calculate reconstruction loss: \quad $\mathcal{L} \gets \|\mathbf{Z}_l - \mathbf{Z}_l'\|_F$
        \STATE back propagation:\\
        \quad \quad $\begin{aligned}
            \frac{\partial\mathcal{L}}{\partial \mathbf{\Sigma}} ~ \text{with detached}~  \mathbf{W}_l ~ \text{and} ~ \mathcal{Q}(\mathbf{W}_l - \mathbf{B}_l \mathbf{A}_l)
        \end{aligned}$
        \STATE update $\mathbf{A}_l,\mathbf{B}_l$ via gradient descent
    \UNTIL{Convergence}
    \STATE $\mathbf{A}_l',\mathbf{B}_l' \gets \mathbf{A}_l,\mathbf{B}_l$
\ENDFOR
\RETURN $\{\mathbf{A}_l', \mathbf{B}_l'\}_{l=1}^{N}$
\end{algorithmic}
\end{algorithm}

\subsection{Kernel Fusion}
\label{sec:kernel}
To address the latency challenge induced by sub-branches describe in \cref{sec:moti-2}, we propose the kernel fusion method, illustrated in \cref{fig:kernel-fusion}. The reconstructed layer consists of four kernels: de-quantization and linear projection in the main path, as well as down-projection and up-projection in the sub-branch.
We observe that while the down-projection requires relatively minimal time, the other operations dominate the overall latency. 
To mitigate this, we integrate weight de-quantization, activation-weight multiplication, and the up-projection into a single CUDA kernel.
This reduces the total number of kernels from four to two, significantly lowering the kernel launch overhead.
Additionally, in the fused kernels, the up-projection in the sub-branch shares the same output tensor as the linear projection in the main path. This optimization minimizes redundant writes, which would otherwise occur if the operations were handled separately.
As a result, these modifications collectively reduce the additional inference delay caused by the sub-branches by 60\%.

\section{Experiments}
In this section, we present the experimental setup of models, baselines, datasets, metrics and implementation details in \cref{sec:ex-setup}. Then, we demonstrate the perplexity and zero-shot accuracy of various quantization methos in \cref{sec:results}, followed by the performance of instruction-tuned models and the wall-clock latency on real devices.
\begin{table*}[ht]
\begin{tabular}{ccc|cccccc}
\hline
Method    & W Bit & Group & Llama2-7B     & Llama2-13B    & Llama2-70B    & Llama3-8B     & Llama3-70B    & Qwen2.5-7B    \\ \hline
FP        & 16    & -     & 5.12          & 4.57          & 3.12          & 5.75          & 2.97          & 6.39          \\ \hline
RTN       & 4     & 128   & 5.27          & 4.69          & 3.25          & 6.29          & 3.63          & 6.72          \\
GPTQ      & 4     & 128   & 5.27          & 4.67          & 3.23          & 7.31          & 3.41          & 6.60          \\
AWQ       & 4     & 128   & 5.23          & 4.65          & 3.20          & 6.11          & 3.25          & 6.62          \\
OmniQuant & 4     & 128   & 5.23          & 4.65          & 3.19          & 6.15          & 3.18          & 6.60          \\
CALDERA   & 4     & 128   & 5.26          & 4.69          & 3.20          & 6.26          & 3.23          & 7.14          \\
SVDQuant  & 4     & 128   & 5.30          & 4.71          & 3.22          & 6.39          & 3.14          & 6.64          \\
\rowcolor[HTML]{F5D7D6} 
\textbf{FBQuant}       & 4     & 128   & \textbf{5.18} & \textbf{4.61} & \textbf{3.15} & \textbf{5.89} & \textbf{3.05} & \textbf{6.52} \\ \hline
RTN       & 3     & 128   & 6.08          & 5.18          & 3.73          & 10.74         & 12.14         & 11.50         \\
GPTQ      & 3     & 128   & 5.96          & 5.07          & 3.69          & 8.66          & 5.04          & 7.90          \\
AWQ       & 3     & 128   & 5.82          & 4.98          & 3.56          & 7.63          & 4.39          & 7.31          \\
OmniQuant & 3     & 128   & 5.78          & 4.96          & 3.53          & 7.86          & 4.12          & 7.23          \\
CALDERA   & 3     & 128   & 5.84          & 5.07          & 3.71          & 9.64          & 4.78          & 10.20         \\
SVDQuant  & 3     & 128   & 6.90          & 5.54          & 3.69          & 10.73         & 4.30          & 8.57          \\
\rowcolor[HTML]{F5D7D6} 
\textbf{FBQuant}       & 3     & 128   & \textbf{5.59} & \textbf{4.86} & \textbf{3.42} & \textbf{6.78} & \textbf{3.77} & \textbf{6.92} \\ \hline
\end{tabular}
\caption{Comparison of perplexity scores on the WikiText2 validation dataset. Lower values indicate better performance. Results for GPTQ, AWQ, and OmniQuant were obtained using their publicly released codebases. Results for CALDERA and SVDQuant were derived from their publicly released codebases with necessary revisions. The best results are highlighted in bold. Our FBQuant approach consistently demonstrates superior performance.}
\label{tab:ppl}
\end{table*}
\begin{table*}[ht]
\begin{tabular}{ccc|cccccc}
\hline
Method    & W Bit & Group & Llama2-7B      & Llama2-13B     & Llama2-70B     & Llama3-8B      & Llama3-70B     & Qwen2.5-7B     \\ \hline
FP        & 16    & -     & 66.62          & 70.45          & 76.27          & 72.79          & 80.22          & 74.25          \\ \hline
RTN       & 4     & 128   & 64.17          & 69.83          & 75.87          & 71.52          & 78.67          & 71.55          \\
GPTQ      & 4     & 128   & 64.78          & 69.93          & 75.85          & 66.12          & 79.51          & 72.09          \\
AWQ       & 4     & 128   & 66.08          & 70.14          & 76.07          & 71.78          & 79.13          & 73.15          \\
OmniQuant & 4     & 128   & 65.75          & 69.85          & 76.10          & 71.81          & 79.37          & 73.41          \\
CALDERA   & 4     & 128   & 65.88          & 68.37          & 76.01          & 70.09          & 79.24          & 68.78          \\
SVDQuant  & 4     & 128   & 65.12          & 69.94          & 76.18          & 71.42          & 77.85          & 73.85          \\
\rowcolor[HTML]{F5D7D6} 
FBQuant   & 4     & 128   & \textbf{66.45} & \textbf{70.05} & \textbf{76.20} & \textbf{72.55} & \textbf{79.74} & \textbf{74.03} \\ \hline
RTN       & 3     & 128   & 62.17          & 68.33          & 73.97          & 61.51          & 68.61          & 67.21          \\
GPTQ      & 3     & 128   & 59.68          & 67.16          & 73.39          & 63.63          & 77.19          & 70.73          \\
AWQ       & 3     & 128   & 63.31          & 68.74          & 74.82          & 67.83          & 77.69          & 70.84          \\
OmniQuant & 3     & 128   & 63.48          & 67.96          & 74.55          & 67.97          & 77.81          & 70.98          \\
CALDERA   & 3     & 128   & 62.10          & 66.13          & 73.56          & 63.48          & 77.17          & 64.45          \\
SVDQuant  & 3     & 128   & 57.06          & 64.49          & 74.61          & 60.38          & 78.47          & 70.79          \\
\rowcolor[HTML]{F5D7D6} 
FBQuant   & 3     & 128   & \textbf{64.68} & \textbf{69.11} & \textbf{75.77} & \textbf{68.87} & \textbf{78.99} & \textbf{72.16} \\ \hline
\end{tabular}
\caption{Comparison of zero-shot accuracy on seven benchmarks as described in the experimental setup. The table shows the average accuracy across all tasks, with the highest score highlighted in bold. Our FBQuant approach consistently demonstrates superior performance. More details are provided in the Appendix.}
\label{tab:0shot}
\end{table*}

\subsection{Experimental Setup}
\label{sec:ex-setup}
\paragraph{Models.} We benchmark our methods using the model frameworks and checkpoints from HuggingFace~\cite{jain2022huggingface}, which includes Llama2~\cite{touvron2023llama}, Llama3~\cite{dubey2024llama}, and Qwen2.5~\cite{qwen2.5} families, with parameter sizes ranging from 7 billion to 70 billion. 
Specifically, we use pre-trained versions to generate the main results as shown in \cref{tab:0shot,tab:ppl}, while instruction-finetuned versions are utilized for the pairwise competition as shown in \cref{fig:win-rate}.
\paragraph{Baselines.} We compare FBQuant against Round-To-Nearest (RTN), GPTQ~\cite{frantar2022gptq}, AWQ~\cite{lin2024awq}, OmniQuant~\cite{shao2023omniquant}, CALDERA~\cite{Saha2024CompressingLL}, and SVDQuant~\cite{li2024svdqunat}. GPTQ, AWQ, and OmniQuant are implemented using their publicly released codebase. For CALDERA and SVDQuant, we follow their papers and codebases to produce the results, as CALDERA only support Lattice quantizer and SVDQuant is originally designed for diffusion models. 
Additionally, we include unquantized models using the float16 datatype as a baseline for a fair comparison.
\paragraph{Datasets and Metrics.} Following previous works~\cite{frantar2022gptq,lin2024awq}, we employ 128 samples with a sequence length of 2048 in the subset of WikiText2~\cite{merity2016pointerwikitext2} training data for calibration. The perplexity results are tested on the WikiText2 validation set. The zero-shot evaluation is conducted using the open-source toolkit, i.e., Language Model Evaluation Harness~\cite{eval-harness}, which has been utilized by other baselines. The evaluation datasets include Arc-Challenge~\cite{allenai:arc}, Arc-Easy~\cite{allenai:arc}, HellaSwag~\cite{zellers2019hellaswag}, MMLU~\cite{hendryckstest2021mmlu}, PIQA~\cite{Bisk2020PIQA}, WinoGrande~\cite{ai2:winogrande}, and BoolQ~\cite{wang2019superglue}. We report the averaged accuracy. 
\paragraph{Implementation Details.} All experiments are conducted using A100 and RTX 3090 GPUs. Both the A100 and 3090 GPUs are utilized for optimizing the sub-branches, while only the 3090 GPU is used for testing latency, as it is commonly available for personal use. In the main results, we set the rank parameter to 128. The total number of optimization epochs is set to 20. A group size of 128 is used in all quantization methods. Sub-branches are integrated into all linear layers in LLMs, such as Query, Key, Value, and Out projections in Attention blocks, as well as Down, Gate, and Up projections in Feed-Forward Networks.
\subsection{Results and Analysis}
\label{sec:results}
\begin{figure}[ht]
    \centering
    \includegraphics[width=\linewidth]{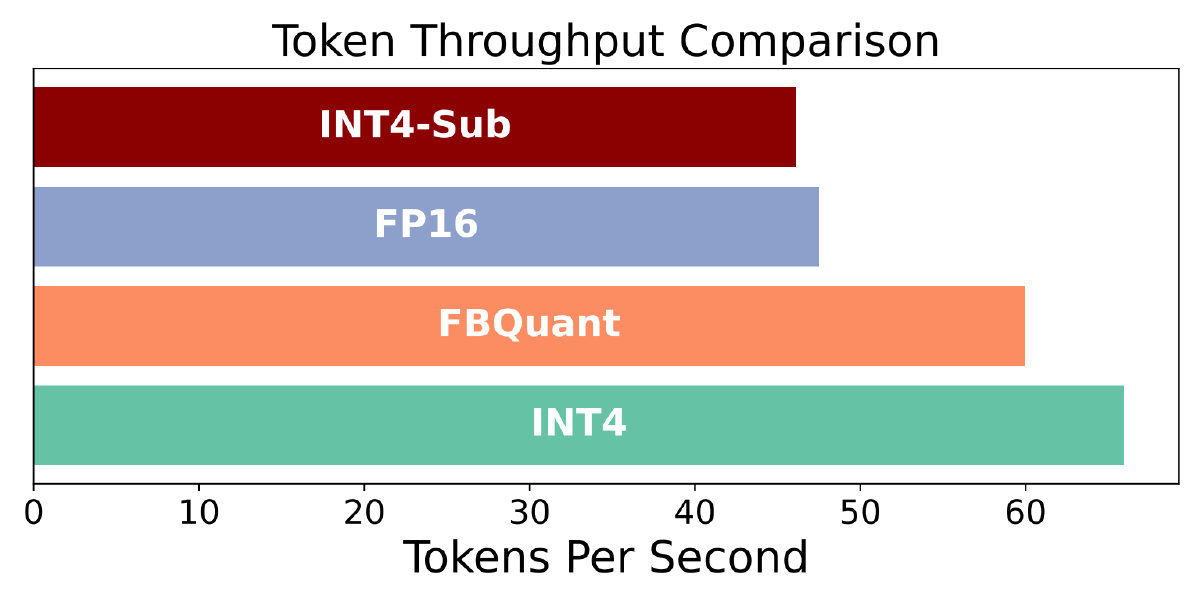}
    \caption{Token throughput of the Llama2-7B model using FP16, INT4-Sub, INT4 and INT4-FBQuant tested on RTX 3090 GPU. Experiments were conducted with a batch size of 1. The rank parameter for INT4-Sub and FBQuant was set to 128. The token throughput is measured by prefilling 256 and decoding 64 tokens.}
    \label{fig:wallclock}
\end{figure}

\paragraph{Performance Comparison.} \cref{tab:ppl} and \cref{tab:0shot} present the evaluation results for perplexity on WikiText2, and zero-shot accuracy on seven public benchmarks, across different quantization methods, tested on various LLM architectures and model sizes. The quantization methods evaluated include RTN, GPTQ, AWQ, and OmniQuant, which rely on clamping and rotation techniques; as well as CALDERA, SVDQuant, and FBQuant, which adopt sub-branch approaches.
Our FBQuant achieves state-of-the-art results in both perplexity and zero-shot accuracy across various models. Specifically, for the 3-bit Llama3-8B model, FBQuant reaches a perplexity of 6.78, marking a significant improvement of 0.85 over the second-best method AWQ. For the Llama2-7B model, FBQuant attains a zero-shot accuracy of 64.68\%, outperforming OmniQuant by 1.20\%.
Three key observations can be made from the results: 
(a) 4-bit quantization generally offers good performance, while 3-bit quantization remains a little gap compared to the floating-point models.
(b) New models present greater challenges for quantization. For instance, the 3-bit Llama3-8B model poses significant challenges for AWQ, which suffers a substantial degradation in perplexity by 1.88.
(c) Existing sub-branch approaches occasionally produce poor performance, as observed with CALDERA on 3-bit version of Llama3-8B, Llama3-70B, and Qwen2.5-7B, as it relies on the ill-posed reconstruction objective as explained in \cref{sec:motivation}. Besides, SVDQuant performs poorly on 3-bit Llama3-8B, as it only focuses on the weight outliers, ignoring the layer output error.
\begin{figure}[t]
    \centering
    \includegraphics[width=\linewidth,]{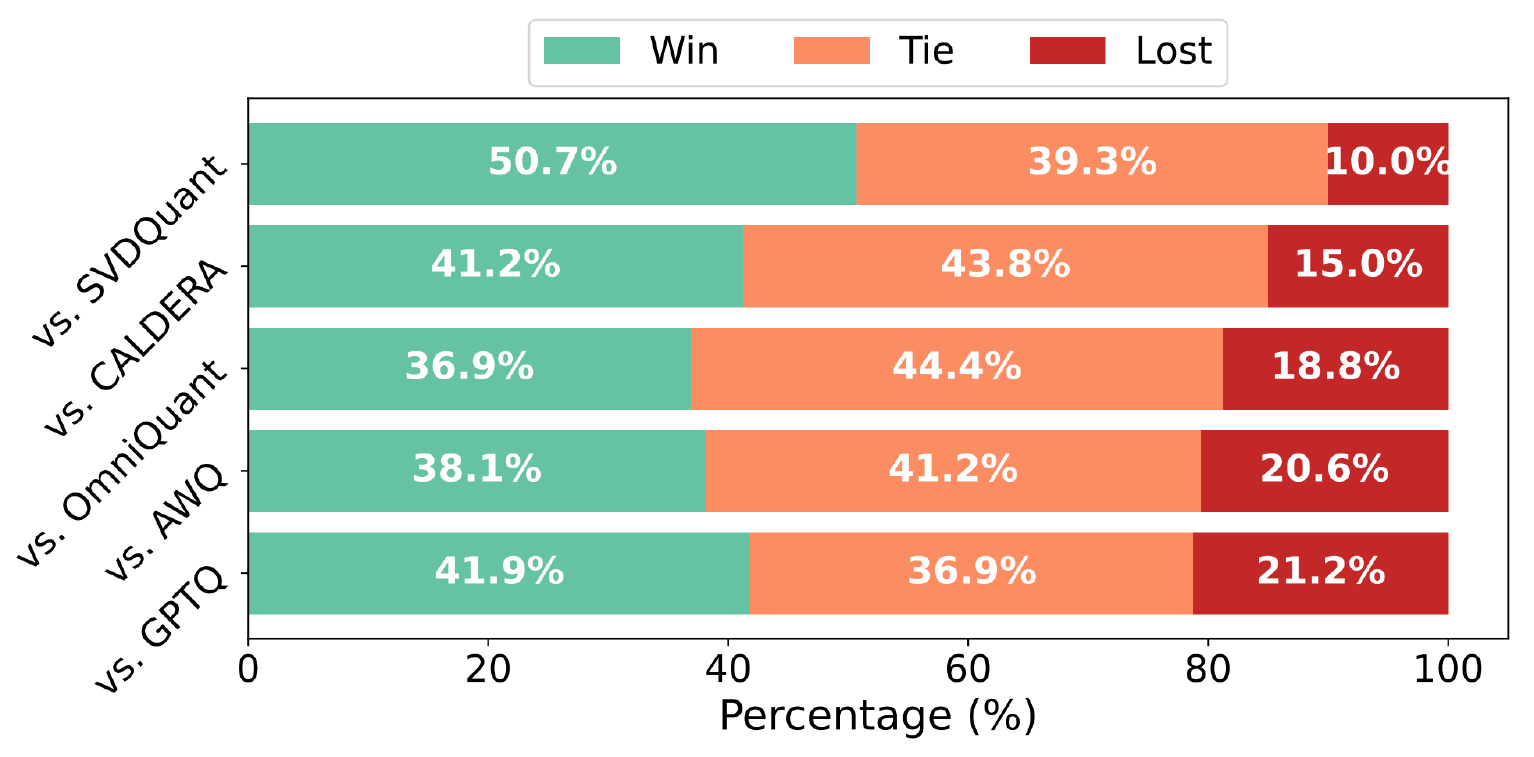}
    \caption{Quantization of Instruction-tuned Models. We use different quantization methods on the Llama3-8B-Chat model to conduct the experiments. FBQuant demonstrates consistent strength against other quantization methods. }
    \label{fig:win-rate}
\end{figure}
\paragraph{Quantization of Instruction-tuned Models.}
Most real-world applications are powered by instruction-tuned models, motivating us to benchmark the performance of our method on the instruction-tuned versions of the corresponding LLMs. Following the experiment setting in \cite{lin2024awq,shao2023omniquant}, we conduct pairwise comparisons among AWQ, OmniQuant, CALDERA, SVDQuant, and FBQuant under 3-bit settings. We use the GPT-4 evaluation protocol~\cite{vicuna2023} to assess performance on the Vicuna benchmark~\cite{vicuna2023}, which comprises 80 questions. The results of the comparisons are reported as win, tie, and loss percentages. A higher percentage of wins and ties indicates better performance, highlighting the superior results achieved by FBQuant. To negate position bias~\cite{zheng2023judging}, we also test pairs in exchanged position, totaling 160 trails per competition. As shown in \cref{fig:win-rate}, in the Llama3-8B-Chat model, our FBQuant demonstrates best performance against other quantization methods. For instance, FBQuant achieves 79.3\% win-tie rate against to AWQ, and 90.0\% win-tie rate against to SVDQuant.
\paragraph{Wall-clock Latency on Real Devices.}
We evaluate the token throughput, measured in tokens per second (tk/s), of the FBQuant kernels on the RTX 3090 GPU, and compare it to floating-point (FP16), INT4, and INT4 with conventional sub-branch (INT4-Sub) implementations. As shown in \cref{fig:wallclock}, INT4-Sub exhibits a significant slowdown, achieving only 46 tk/s, slightly slower than FP16, which achieves 48 tk/s. In contrast, FBQuant demonstrates a substantial improvement against baselines, achieving 61 tk/s and significantly increasing token throughput.

\section{Conclusions}
In this paper, we introduced FBQuant, a novel feedback-based optimization approach that addresses key challenges in the sub-branching quantization for LLMs. By incorporating a feedback mechanism inspired by automatic control systems, FBQuant effectively optimizes sub-branches without overfitting to calibration data, ensuring robust reconstruction of quantized weights. Additionally, we developed an efficient CUDA kernel fusion implementation to tackle the latency overhead introduced by sub-branches, significantly reducing inference delays caused by memory access bottlenecks.
Our experiments demonstrated that FBQuant consistently outperforms existing quantization techniques across various tasks and model families on LLMs, achieving superior accuracy and perplexity while improving inference efficiency. Specifically, FBQuant enhances zero-shot accuracy for 3-bit models, such as Llama2-7B, by 1.2\%, and achieves significant throughput improvements in wall-clock evaluations. These results highlight the practical utility of FBQuant for on-device deployment of LLMs, where computational and memory constraints are critical.


\section*{Acknowledgments}
This work was supported in part by the National Key Research and Development Program of China under Grant No. 2022YFB4400900, in part by the Strategic Industries and Key Technologies Project of Jiangsu Province under Grant BE2023020-3 and the Basic Research Program of Jiangsu Province under Grant BK20243042.

\bibliographystyle{named}
\bibliography{ijcai25}

\begin{thebibliography}{}

\bibitem[\protect\citeauthoryear{Ashkboos \bgroup \em et al.\egroup }{2024}]{ashkboos2024quarot}
Saleh Ashkboos, Amirkeivan Mohtashami, Maximilian~L Croci, Bo~Li, Pashmina Cameron, Martin Jaggi, Dan Alistarh, Torsten Hoefler, and James Hensman.
\newblock Quarot: Outlier-free 4-bit inference in rotated llms.
\newblock {\em arXiv preprint arXiv:2404.00456}, 2024.

\bibitem[\protect\citeauthoryear{Bengio \bgroup \em et al.\egroup }{2013}]{bengio2013estimatingSTE}
Yoshua Bengio, Nicholas L{\'e}onard, and Aaron Courville.
\newblock Estimating or propagating gradients through stochastic neurons for conditional computation.
\newblock {\em arXiv preprint arXiv:1308.3432}, 2013.

\bibitem[\protect\citeauthoryear{Bisk \bgroup \em et al.\egroup }{2020}]{Bisk2020PIQA}
Yonatan Bisk, Rowan Zellers, Ronan~Le Bras, Jianfeng Gao, and Yejin Choi.
\newblock Piqa: Reasoning about physical commonsense in natural language.
\newblock In {\em Thirty-Fourth AAAI Conference on Artificial Intelligence}, 2020.

\bibitem[\protect\citeauthoryear{Chiang \bgroup \em et al.\egroup }{2023}]{vicuna2023}
Wei-Lin Chiang, Zhuohan Li, Zi~Lin, Ying Sheng, Zhanghao Wu, Hao Zhang, Lianmin Zheng, Siyuan Zhuang, Yonghao Zhuang, Joseph~E. Gonzalez, Ion Stoica, and Eric~P. Xing.
\newblock Vicuna: An open-source chatbot impressing gpt-4 with 90\%* chatgpt quality, March 2023.

\bibitem[\protect\citeauthoryear{Clark \bgroup \em et al.\egroup }{2018}]{allenai:arc}
Peter Clark, Isaac Cowhey, Oren Etzioni, Tushar Khot, Ashish Sabharwal, Carissa Schoenick, and Oyvind Tafjord.
\newblock Think you have solved question answering? try arc, the ai2 reasoning challenge.
\newblock {\em arXiv:1803.05457v1}, 2018.

\bibitem[\protect\citeauthoryear{Dettmers \bgroup \em et al.\egroup }{2022}]{dettmers2022gpt3}
Tim Dettmers, Mike Lewis, Younes Belkada, and Luke Zettlemoyer.
\newblock Gpt3. int8 (): 8-bit matrix multiplication for transformers at scale.
\newblock {\em Advances in Neural Information Processing Systems}, 35:30318--30332, 2022.

\bibitem[\protect\citeauthoryear{Dettmers \bgroup \em et al.\egroup }{2024}]{dettmers2024qlora}
Tim Dettmers, Artidoro Pagnoni, Ari Holtzman, and Luke Zettlemoyer.
\newblock Qlora: Efficient finetuning of quantized llms.
\newblock {\em Advances in Neural Information Processing Systems}, 36, 2024.

\bibitem[\protect\citeauthoryear{Dong \bgroup \em et al.\egroup }{2019}]{dong2019hawq}
Zhen Dong, Zhewei Yao, Amir Gholami, Michael~W Mahoney, and Kurt Keutzer.
\newblock Hawq: Hessian aware quantization of neural networks with mixed-precision.
\newblock In {\em Proceedings of the IEEE/CVF international conference on computer vision}, pages 293--302, 2019.

\bibitem[\protect\citeauthoryear{Dubey \bgroup \em et al.\egroup }{2024}]{dubey2024llama}
Abhimanyu Dubey, Abhinav Jauhri, Abhinav Pandey, Abhishek Kadian, Ahmad Al-Dahle, Aiesha Letman, Akhil Mathur, Alan Schelten, Amy Yang, Angela Fan, et~al.
\newblock The llama 3 herd of models.
\newblock {\em arXiv preprint arXiv:2407.21783}, 2024.

\bibitem[\protect\citeauthoryear{Franklin \bgroup \em et al.\egroup }{2002}]{franklin2002feedback}
Gene~F Franklin, J~David Powell, Abbas Emami-Naeini, and J~David Powell.
\newblock {\em Feedback control of dynamic systems}, volume~4.
\newblock Prentice hall Upper Saddle River, 2002.

\bibitem[\protect\citeauthoryear{Frantar and Alistarh}{2022}]{frantar2022optimal}
Elias Frantar and Dan Alistarh.
\newblock Optimal brain compression: A framework for accurate post-training quantization and pruning.
\newblock {\em Advances in Neural Information Processing Systems}, 35:4475--4488, 2022.

\bibitem[\protect\citeauthoryear{Frantar \bgroup \em et al.\egroup }{2022}]{frantar2022gptq}
Elias Frantar, Saleh Ashkboos, Torsten Hoefler, and Dan Alistarh.
\newblock Gptq: Accurate post-training quantization for generative pre-trained transformers.
\newblock {\em arXiv preprint arXiv:2210.17323}, 2022.

\bibitem[\protect\citeauthoryear{Gao \bgroup \em et al.\egroup }{2024}]{eval-harness}
Leo Gao, Jonathan Tow, Baber Abbasi, Stella Biderman, Sid Black, Anthony DiPofi, Charles Foster, Laurence Golding, Jeffrey Hsu, Alain Le~Noac'h, Haonan Li, Kyle McDonell, Niklas Muennighoff, Chris Ociepa, Jason Phang, Laria Reynolds, Hailey Schoelkopf, Aviya Skowron, Lintang Sutawika, Eric Tang, Anish Thite, Ben Wang, Kevin Wang, and Andy Zou.
\newblock A framework for few-shot language model evaluation, 07 2024.

\bibitem[\protect\citeauthoryear{Hendrycks \bgroup \em et al.\egroup }{2021}]{hendryckstest2021mmlu}
Dan Hendrycks, Collin Burns, Steven Basart, Andy Zou, Mantas Mazeika, Dawn Song, and Jacob Steinhardt.
\newblock Measuring massive multitask language understanding.
\newblock {\em Proceedings of the International Conference on Learning Representations (ICLR)}, 2021.

\bibitem[\protect\citeauthoryear{Hu \bgroup \em et al.\egroup }{2021}]{hu2021lora}
Edward~J Hu, Yelong Shen, Phillip Wallis, Zeyuan Allen-Zhu, Yuanzhi Li, Shean Wang, Lu~Wang, and Weizhu Chen.
\newblock Lora: Low-rank adaptation of large language models.
\newblock {\em arXiv preprint arXiv:2106.09685}, 2021.

\bibitem[\protect\citeauthoryear{Huang \bgroup \em et al.\egroup }{2024}]{huang2024billm}
Wei Huang, Yangdong Liu, Haotong Qin, Ying Li, Shiming Zhang, Xianglong Liu, Michele Magno, and Xiaojuan Qi.
\newblock Billm: Pushing the limit of post-training quantization for llms.
\newblock {\em arXiv preprint arXiv:2402.04291}, 2024.

\bibitem[\protect\citeauthoryear{Jain}{2022}]{jain2022huggingface}
Shashank~Mohan Jain.
\newblock Hugging face.
\newblock In {\em Introduction to transformers for NLP: With the hugging face library and models to solve problems}, pages 51--67. Springer, 2022.

\bibitem[\protect\citeauthoryear{Li \bgroup \em et al.\egroup }{2023a}]{li2023loftq}
Yixiao Li, Yifan Yu, Chen Liang, Pengcheng He, Nikos Karampatziakis, Weizhu Chen, and Tuo Zhao.
\newblock Loftq: Lora-fine-tuning-aware quantization for large language models.
\newblock {\em arXiv preprint arXiv:2310.08659}, 2023.

\bibitem[\protect\citeauthoryear{Li \bgroup \em et al.\egroup }{2023b}]{Li2023MetaAgentsSI}
Yuan Li, Yixuan Zhang, and Lichao Sun.
\newblock Metaagents: Simulating interactions of human behaviors for llm-based task-oriented coordination via collaborative generative agents.
\newblock {\em ArXiv}, abs/2310.06500, 2023.

\bibitem[\protect\citeauthoryear{Li \bgroup \em et al.\egroup }{2024}]{li2024svdqunat}
Muyang Li, Yujun Lin, Zhekai Zhang, Tianle Cai, Xiuyu Li, Junxian Guo, Enze Xie, Chenlin Meng, Jun-Yan Zhu, and Song Han.
\newblock Svdqunat: Absorbing outliers by low-rank components for 4-bit diffusion models.
\newblock {\em arXiv preprint arXiv:2411.05007}, 2024.

\bibitem[\protect\citeauthoryear{Lin \bgroup \em et al.\egroup }{2024a}]{lin2024duquant}
Haokun Lin, Haobo Xu, Yichen Wu, Jingzhi Cui, Yingtao Zhang, Linzhan Mou, Linqi Song, Zhenan Sun, and Ying Wei.
\newblock Duquant: Distributing outliers via dual transformation makes stronger quantized llms.
\newblock In {\em The Thirty-eighth Annual Conference on Neural Information Processing Systems}, 2024.

\bibitem[\protect\citeauthoryear{Lin \bgroup \em et al.\egroup }{2024b}]{lin2024awq}
Ji~Lin, Jiaming Tang, Haotian Tang, Shang Yang, Wei-Ming Chen, Wei-Chen Wang, Guangxuan Xiao, Xingyu Dang, Chuang Gan, and Song Han.
\newblock Awq: Activation-aware weight quantization for on-device llm compression and acceleration.
\newblock {\em Proceedings of Machine Learning and Systems}, 6:87--100, 2024.

\bibitem[\protect\citeauthoryear{Lin \bgroup \em et al.\egroup }{2024c}]{lin2024qserve}
Yujun Lin, Haotian Tang, Shang Yang, Zhekai Zhang, Guangxuan Xiao, Chuang Gan, and Song Han.
\newblock Qserve: W4a8kv4 quantization and system co-design for efficient llm serving.
\newblock {\em arXiv preprint arXiv:2405.04532}, 2024.

\bibitem[\protect\citeauthoryear{Liu \bgroup \em et al.\egroup }{2023a}]{liu2023noisyquant}
Yijiang Liu, Huanrui Yang, Zhen Dong, Kurt Keutzer, Li~Du, and Shanghang Zhang.
\newblock Noisyquant: Noisy bias-enhanced post-training activation quantization for vision transformers.
\newblock In {\em Proceedings of the IEEE/CVF Conference on Computer Vision and Pattern Recognition}, pages 20321--20330, 2023.

\bibitem[\protect\citeauthoryear{Liu \bgroup \em et al.\egroup }{2023b}]{Liu2023LLMQATDQ}
Zechun Liu, Barlas Oğuz, Changsheng Zhao, Ernie Chang, Pierre Stock, Yashar Mehdad, Yangyang Shi, Raghuraman Krishnamoorthi, and Vikas Chandra.
\newblock Llm-qat: Data-free quantization aware training for large language models.
\newblock {\em ArXiv}, abs/2305.17888, 2023.

\bibitem[\protect\citeauthoryear{Liu \bgroup \em et al.\egroup }{2024a}]{liu2024eora}
Shih-Yang Liu, Huck Yang, Chein-Yi Wang, Nai~Chit Fung, Hongxu Yin, Charbel Sakr, Saurav Muralidharan, Kwang-Ting Cheng, Jan Kautz, Yu-Chiang~Frank Wang, et~al.
\newblock Eora: Training-free compensation for compressed llm with eigenspace low-rank approximation.
\newblock {\em arXiv preprint arXiv:2410.21271}, 2024.

\bibitem[\protect\citeauthoryear{Liu \bgroup \em et al.\egroup }{2024b}]{liu2024spinquant}
Zechun Liu, Changsheng Zhao, Igor Fedorov, Bilge Soran, Dhruv Choudhary, Raghuraman Krishnamoorthi, Vikas Chandra, Yuandong Tian, and Tijmen Blankevoort.
\newblock Spinquant--llm quantization with learned rotations.
\newblock {\em arXiv preprint arXiv:2405.16406}, 2024.

\bibitem[\protect\citeauthoryear{Merity \bgroup \em et al.\egroup }{2016}]{merity2016pointerwikitext2}
Stephen Merity, Caiming Xiong, James Bradbury, and Richard Socher.
\newblock Pointer sentinel mixture models, 2016.

\bibitem[\protect\citeauthoryear{Nagel \bgroup \em et al.\egroup }{2020}]{Nagel2020adaround}
Markus Nagel, Rana~Ali Amjad, Mart van Baalen, Christos Louizos, and Tijmen Blankevoort.
\newblock Up or down? adaptive rounding for post-training quantization.
\newblock {\em ArXiv}, abs/2004.10568, 2020.

\bibitem[\protect\citeauthoryear{Saha \bgroup \em et al.\egroup }{2024}]{Saha2024CompressingLL}
Rajarshi Saha, Naomi Sagan, Varun Srivastava, Andrea~J. Goldsmith, and Mert Pilanci.
\newblock Compressing large language models using low rank and low precision decomposition.
\newblock {\em ArXiv}, abs/2405.18886, 2024.

\bibitem[\protect\citeauthoryear{Sakaguchi \bgroup \em et al.\egroup }{2019}]{ai2:winogrande}
Keisuke Sakaguchi, Ronan~Le Bras, Chandra Bhagavatula, and Yejin Choi.
\newblock An adversarial winograd schema challenge at scale.
\newblock 2019.

\bibitem[\protect\citeauthoryear{Shao \bgroup \em et al.\egroup }{2023}]{shao2023omniquant}
Wenqi Shao, Mengzhao Chen, Zhaoyang Zhang, Peng Xu, Lirui Zhao, Zhiqian Li, Kaipeng Zhang, Peng Gao, Yu~Qiao, and Ping Luo.
\newblock Omniquant: Omnidirectionally calibrated quantization for large language models.
\newblock {\em arXiv preprint arXiv:2308.13137}, 2023.

\bibitem[\protect\citeauthoryear{Talebirad and Nadiri}{2023}]{Talebirad2023MultiAgentCH}
Yashar Talebirad and Amirhossein Nadiri.
\newblock Multi-agent collaboration: Harnessing the power of intelligent llm agents.
\newblock {\em ArXiv}, abs/2306.03314, 2023.

\bibitem[\protect\citeauthoryear{Touvron \bgroup \em et al.\egroup }{2023}]{touvron2023llama}
Hugo Touvron, Thibaut Lavril, Gautier Izacard, Xavier Martinet, Marie-Anne Lachaux, Timoth{\'e}e Lacroix, Baptiste Rozi{\`e}re, Naman Goyal, Eric Hambro, Faisal Azhar, et~al.
\newblock Llama: Open and efficient foundation language models.
\newblock {\em arXiv preprint arXiv:2302.13971}, 2023.

\bibitem[\protect\citeauthoryear{Wang \bgroup \em et al.\egroup }{2019}]{wang2019superglue}
Alex Wang, Yada Pruksachatkun, Nikita Nangia, Amanpreet Singh, Julian Michael, Felix Hill, Omer Levy, and Samuel~R. Bowman.
\newblock Super{GLUE}: A stickier benchmark for general-purpose language understanding systems.
\newblock {\em arXiv preprint 1905.00537}, 2019.

\bibitem[\protect\citeauthoryear{Wei \bgroup \em et al.\egroup }{2022}]{Wei2022OutlierSP}
Xiuying Wei, Yunchen Zhang, Xiangguo Zhang, Ruihao Gong, Shanghang Zhang, Qi~Zhang, Fengwei Yu, and Xianglong Liu.
\newblock Outlier suppression: Pushing the limit of low-bit transformer language models.
\newblock {\em ArXiv}, abs/2209.13325, 2022.

\bibitem[\protect\citeauthoryear{Xiao \bgroup \em et al.\egroup }{2022}]{Xiao2022SmoothQuantAA}
Guangxuan Xiao, Ji~Lin, Mickael Seznec, Julien Demouth, and Song Han.
\newblock Smoothquant: Accurate and efficient post-training quantization for large language models.
\newblock {\em ArXiv}, abs/2211.10438, 2022.

\bibitem[\protect\citeauthoryear{Yang \bgroup \em et al.\egroup }{2024}]{qwen2.5}
An~Yang, Baosong Yang, Beichen Zhang, Binyuan Hui, Bo~Zheng, Bowen Yu, Chengyuan Li, Dayiheng Liu, Fei Huang, Haoran Wei, Huan Lin, Jian Yang, Jianhong Tu, Jianwei Zhang, Jianxin Yang, Jiaxi Yang, Jingren Zhou, Junyang Lin, Kai Dang, Keming Lu, Keqin Bao, Kexin Yang, Le~Yu, Mei Li, Mingfeng Xue, Pei Zhang, Qin Zhu, Rui Men, Runji Lin, Tianhao Li, Tingyu Xia, Xingzhang Ren, Xuancheng Ren, Yang Fan, Yang Su, Yichang Zhang, Yu~Wan, Yuqiong Liu, Zeyu Cui, Zhenru Zhang, and Zihan Qiu.
\newblock Qwen2.5 technical report.
\newblock {\em arXiv preprint arXiv:2412.15115}, 2024.

\bibitem[\protect\citeauthoryear{Yuan \bgroup \em et al.\egroup }{2023}]{Yuan2023RPTQRP}
Zhihang Yuan, Lin Niu, Jia-Wen Liu, Wenyu Liu, Xinggang Wang, Yuzhang Shang, Guangyu Sun, Qiang Wu, Jiaxiang Wu, and Bingzhe Wu.
\newblock Rptq: Reorder-based post-training quantization for large language models.
\newblock {\em ArXiv}, abs/2304.01089, 2023.

\bibitem[\protect\citeauthoryear{Zellers \bgroup \em et al.\egroup }{2019}]{zellers2019hellaswag}
Rowan Zellers, Ari Holtzman, Yonatan Bisk, Ali Farhadi, and Yejin Choi.
\newblock Hellaswag: Can a machine really finish your sentence?
\newblock In {\em Proceedings of the 57th Annual Meeting of the Association for Computational Linguistics}, 2019.

\bibitem[\protect\citeauthoryear{Zheng \bgroup \em et al.\egroup }{2023}]{zheng2023judging}
Lianmin Zheng, Wei-Lin Chiang, Ying Sheng, Siyuan Zhuang, Zhanghao Wu, Yonghao Zhuang, Zi~Lin, Zhuohan Li, Dacheng Li, Eric Xing, et~al.
\newblock Judging llm-as-a-judge with mt-bench and chatbot arena.
\newblock {\em Advances in Neural Information Processing Systems}, 36:46595--46623, 2023.

\end{thebibliography}

\newpage
\appendix
\section{Detailed Ill-posed Optimization}
The sub-branch is often implemented by the LoRA~\cite{hu2021lora} approach using down and up projections,
\begin{equation}
    \mathbf{\Sigma} = \mathbf{B\cdot A},
\end{equation}
where $\mathbf{\Sigma}$ is the equivalent weights of the sub-branch.
Then, sub-branch compensation reconstructs the weights of a layer as:
\begin{equation}
\begin{aligned}
    \mathbf{W}' &= \mathbf{W}_\mathcal{Q} + \mathbf{B\cdot A}\\
    &= \mathbf{W}_\mathcal{Q} + \mathbf{\Sigma},
\end{aligned}
\end{equation}
where $\mathbf{W}$  is the original weight matrix, $\mathbf{W}_\mathcal{Q}$ is its quantized counterpart produced by quantizer $\mathcal{Q}$.
Then, the layer output derived by the original layer and the reconstructed layer are, respectively, $\textbf{Z}$ and $\textbf{Z}'$:
\begin{equation}
\begin{aligned}
    \textbf{Z} &= \textbf{WX}^{\top},\\
    \textbf{Z}' &= \textbf{W}' \cdot \textbf{X}^{\top},
\end{aligned}
\end{equation}
To optimize the sub-branch layer by layer, we consider the layer-wise reconstruction loss as minimizing the output error:
\begin{equation}
\begin{aligned}
\mathbf{\Sigma}^* = \arg\min_{\mathbf{\Sigma}}\mathcal{L}_1,& \quad \text{subject to} \quad \text{rank}(\mathbf{\Sigma}) \leq r,
\end{aligned}
\end{equation}
where
\begin{equation}
\begin{aligned}
    \mathcal{L}_1 &= \|\mathbf{Z}-\mathbf{Z}'\|_F\\
    &= \|(\mathbf{W} - \mathbf{W}')\mathbf{X}^\top \|_F\\
    &= \|(\mathbf{W} - \mathbf{W}_{\mathcal{Q}} - \mathbf{\Sigma})\mathbf{X}^\top \|_F,
\end{aligned}
\end{equation}
$\|\cdot\|_F$ denotes the Frobenius norm, $\mathbf{X}$ represents layer inputs, and $r$ is the rank constraint for the sub-branch. Letting  $\mathbf{\Delta} = \mathbf{W}-\mathbf{W}_\mathcal{Q}$, assume $\mathbf{\Sigma}^*$ yields a minimal value $\epsilon_1$ for $\mathcal{L}_1$, then 
\begin{equation}
\begin{aligned}
    \epsilon_1 &= \|(\mathbf{\Delta} -\mathbf{\Sigma}^*) \mathbf{X}^\top\|_F\\
    &=\text{tr} \left( (\mathbf{\Delta} -\mathbf{\Sigma}^*) \mathbf{X}^\top \mathbf{X} (\mathbf{\Delta} -\mathbf{\Sigma}^*)^\top \right),
\end{aligned}
\end{equation}
where $\text{tr}(\cdot)$ denotes the trace operation.
In addition, from the Singular Value Decomposition (SVD), $\mathbf{\Sigma}^*$ can be expanded in terms of the top-$r$ singular vectors:
\begin{equation}
    \mathbf{\Sigma}^*=\mathbf{U}_r\mathbf{S}_r\mathbf{V}_r.
\end{equation}
We can subsequently consider an alternative solution $\mathbf{\Sigma}'$:
\begin{equation}
    \mathbf{\Sigma}'=\mathbf{\Sigma}^* + \mathbf{\Sigma}_N,\quad \text{where}\quad \mathbf{\Sigma}_N = \mathbf{U}_r\mathbf{S}_r(\alpha\mathbf{N}_r),
\end{equation}
with $\mathbf{N}_r$ sharing the same dimensionality as $\mathbf{V}_r$, and a scalar $\alpha$. Because typical calibration data are limited, $\mathbf{X}^\top\mathbf{X}$ is positive semidefinite but not strictly full-rank~\cite{huang2024billm}, implying there exists non-zero $\mathbf{N}_r$ orthogonal to $\mathbf{X}^\top\mathbf{X}$, that is,
\begin{equation}
    \mathbf{N}_r\mathbf{X}^\top \mathbf{X} = \mathbf{0}.
\end{equation}
Then,
\begin{equation}
\begin{aligned}
    \mathbf{\Sigma}_N\mathbf{X}^\top \mathbf{X} &=\mathbf{U}_r\mathbf{S}_r(\alpha\mathbf{N}_r\mathbf{X}^\top \mathbf{X})\\
    &= \mathbf{0}.
\end{aligned}
\end{equation}
Hence, the construction loss $\epsilon'$ achieved by $\mathbf{\Sigma}'$ is equal to $\epsilon_1$:
\begin{equation}
\begin{aligned}
    \epsilon' &= \text{tr} \left( (\mathbf{\Delta} -\mathbf{\Sigma}') \mathbf{X}^\top \mathbf{X} (\mathbf{\Delta} -\mathbf{\Sigma}')^\top \right)\\
    &= \text{tr} \left( (\mathbf{\Delta} -\mathbf{\Sigma}^*-\mathbf{\Sigma}_N) \mathbf{X}^\top \mathbf{X} (\mathbf{\Delta} -\mathbf{\Sigma}^* -\mathbf{\Sigma}_N)^\top \right)\\
    &= \text{tr} \left( (\mathbf{\Delta} -\mathbf{\Sigma}^*) \mathbf{X}^\top \mathbf{X} (\mathbf{\Delta} -\mathbf{\Sigma}^*)^\top \right) \\
    &\quad \quad  \color{gray}-\text{tr} \left(  (\mathbf{\Delta}-\mathbf{\Sigma}^*)(\mathbf{\Sigma}_N \mathbf{X}^\top \mathbf{X} )^\top \right) \\
    &\quad \quad  \color{gray}-\text{tr} \left( (\mathbf{\Sigma}_N \mathbf{X}^\top \mathbf{X}) \cdot (\mathbf{\Delta} -\mathbf{\Sigma}^*-\mathbf{\Sigma}_N)^\top \right) \\
    &= \text{tr} \left( (\mathbf{\Delta} -\mathbf{\Sigma}^*) \mathbf{X}^\top \mathbf{X} (\mathbf{\Delta} -\mathbf{\Sigma}^*)^\top \right) \\
    &= \epsilon_1,
\end{aligned}
\end{equation}
where gray terms are $\mathbf{0}$ due to orthogonality. 
This equation demonstrates that $\mathbf{\Sigma}'$ is also a valid solution.
However, for any value $w\in \mathbf{W}$, which is reconstructed by $\sigma' \in \mathbf{\Sigma}'$, the reconstructed value is:
\begin{equation}
\begin{aligned}
w'&=\mathcal{Q}(w)+\sigma'\\
&=  \mathcal{Q}(w)+\sigma^* + \alpha\sigma_N,
\end{aligned}
\end{equation}
then,
the difference between the original and the reconstructed weights is:
\begin{equation}
\begin{aligned}
    |w - w'| &= \big|w-\big(\mathcal{Q}(w)+\sigma'\big)\big| \\
    &=|w - \mathcal{Q}(w) - \sigma^* - \alpha\sigma_N|.
\end{aligned}
\end{equation}
With determined $\{w,\mathcal{Q},\sigma^*\}$, the unbounded term $\alpha\sigma_N$ may significantly deviate the reconstructed weights, leading to less meaningful values, which highlights the potential for overfitting. 

\section{Detailed Zero-shot Accuracy Results}
We give detailed zero-shot results in \cref{tab:ap1,tab:ap2,tab:ap3,tab:ap4,tab:ap5,tab:ap6}. The averaged accuracy is highlighted in bold.

\begin{table*}[ht]
\caption{Zero-shot accuracy of Llama2-7B.}
\label{tab:ap1}
\begin{tabular}{@{}ccc|cccccccc@{}}
\toprule
\textbf{Method} & \textbf{Wbit} & \textbf{Group} & \textbf{Avg.}  & \textbf{Arc-c} & \textbf{Arc-e} & \textbf{HellaSwag} & \textbf{MMLU} & \textbf{PIQA} & \textbf{WinoGrande} & \textbf{BoolQ} \\ \midrule
FP                 & 16            & -              & \textbf{66.62} & 46.25          & 76.35          & 76.00              & 41.84         & 79.11         & 69.06               & 77.71          \\ \midrule
RTN                & 4             & 128            & \textbf{64.17} & 42.22          & 74.22          & 74.36              & 39.45         & 77.60         & 66.82               & 74.55          \\
GPTQ               & 4             & 128            & \textbf{64.78} & 43.09          & 75.00          & 74.76              & 38.73         & 78.45         & 68.27               & 75.14          \\
AWQ                & 4             & 128            & \textbf{66.08} & 45.31          & 75.59          & 75.57              & 40.75         & 78.78         & 68.35               & 78.23          \\
OmniQuant          & 4             & 128            & \textbf{65.75} & 44.54          & 75.21          & 74.88              & 41.35         & 78.94         & 68.19               & 77.16          \\
CALDERA            & 4             & 128            & \textbf{65.88} & 43.84          & 75.70          & 75.79              & 39.41         & 81.02         & 68.45               & 76.96          \\
SVDQuant           & 4             & 128            & \textbf{65.12} & 43.86          & 74.92          & 75.55              & 37.29         & 78.40         & 69.69               & 76.15          \\
\rowcolor[HTML]{F5D7D6} 
FBQuant            & 4             & 128            & \textbf{66.45} & 45.37          & 75.84          & 75.30              & 41.42         & 79.11         & 69.64               & 78.50          \\ \midrule
RTN                & 3             & 128            & \textbf{62.17} & 41.47          & 73.19          & 72.30              & 33.51         & 76.22         & 66.93               & 71.56          \\
GPTQ               & 3             & 128            & \textbf{59.68} & 38.91          & 70.54          & 70.10              & 32.94         & 76.78         & 67.48               & 61.01          \\
AWQ                & 3             & 128            & \textbf{63.31} & 44.37          & 73.91          & 73.33              & 32.75         & 77.26         & 68.19               & 73.36          \\
OmniQuant          & 3             & 128            & \textbf{63.48} & 42.24          & 74.87          & 73.03              & 36.05         & 78.07         & 67.88               & 72.23          \\
CALDERA            & 3             & 128            & \textbf{62.10} & 41.25          & 73.01          & 69.98              & 34.87         & 77.98         & 66.59               & 71.03          \\
SVDQuant           & 3             & 128            & \textbf{57.06} & 40.27          & 69.74          & 69.43              & 25.54         & 76.12         & 64.80               & 53.52          \\
\rowcolor[HTML]{F5D7D6} 
FBQuant            & 3             & 128            & \textbf{64.68} & 44.28          & 73.57          & 74.87              & 38.18         & 77.37         & 68.11               & 76.36          \\ \bottomrule
\end{tabular}
\end{table*}

\begin{table*}[ht]
\caption{Zero-shot accuracy of Llama2-13B.}
\label{tab:ap2}
\begin{tabular}{@{}ccc|cccccccc@{}}
\toprule
\textbf{Method} & \textbf{Wbit} & \textbf{Group} & \multicolumn{1}{c}{\textbf{Avg.}} & \multicolumn{1}{c}{\textbf{Arc-c}} & \multicolumn{1}{c}{\textbf{Arc-e}} & \multicolumn{1}{c}{\textbf{HellaSwag}} & \multicolumn{1}{c}{\textbf{MMLU}} & \multicolumn{1}{c}{\textbf{PIQA}} & \multicolumn{1}{c}{\textbf{WinoGrande}} & \multicolumn{1}{c}{\textbf{BoolQ}} \\ \midrule
FP              & 16            & -              & \textbf{70.45}                    & 49.06                              & 79.42                              & 79.39                                  & 52.04                             & 80.52                             & 72.14                                   & 80.58                              \\ \midrule
RTN             & 4             & 128            & \textbf{69.83}                    & 48.12                              & 78.75                              & 78.72                                  & 50.90                             & 80.41                             & 71.98                                   & 79.90                              \\
GPTQ            & 4             & 128            & \textbf{69.93}                    & 48.89                              & 79.25                              & 78.71                                  & 51.18                             & 80.03                             & 71.59                                   & 79.89                              \\
AWQ             & 4             & 128            & \textbf{70.14}                    & 48.21                              & 79.29                              & 78.85                                  & 51.08                             & 80.09                             & 73.40                                   & 80.03                              \\
OmniQuant       & 4             & 128            & \textbf{69.85}                    & 49.15                              & 78.16                              & 78.30                                  & 51.55                             & 79.87                             & 72.45                                   & 79.47                              \\
CALDERA         & 4             & 128            & \textbf{68.37}                    & 48.76                              & 78.67                              & 77.82                                  & 50.80                             & 76.46                             & 70.59                                   & 75.45                              \\
SVDQuant        & 4             & 128            & \textbf{69.94}                    & 49.74                              & 79.88                              & 78.73                                  & 49.81                             & 80.14                             & 72.06                                   & 79.24                              \\
\rowcolor[HTML]{F5D7D6} 
FBQuant         & 4             & 128            & \textbf{70.05}                    & 49.74                              & 79.26                              & 79.01                                  & 49.91                             & 80.44                             & 73.83                                   & 78.15                              \\ \midrule
RTN             & 3             & 128            & \textbf{68.33}                    & 49.15                              & 77.40                              & 76.68                                  & 46.97                             & 79.43                             & 70.88                                   & 77.83                              \\
GPTQ            & 3             & 128            & \textbf{67.16}                    & 45.22                              & 76.26                              & 75.66                                  & 46.78                             & 78.45                             & 70.96                                   & 76.79                              \\
AWQ             & 3             & 128            & \textbf{68.74}                    & 48.81                              & 77.78                              & 77.53                                  & 48.98                             & 78.94                             & 72.45                                   & 76.69                              \\
OmniQuant       & 3             & 128            & \textbf{67.96}                    & 47.01                              & 77.61                              & 76.58                                  & 47.27                             & 79.43                             & 70.09                                   & 77.73                              \\
CALDERA         & 3             & 128            & \textbf{66.13}                    & 47.06                              & 73.29                              & 72.60                                  & 46.43                             & 78.21                             & 69.60                                   & 75.74                              \\
SVDQuant        & 3             & 128            & \textbf{64.49}                    & 43.43                              & 74.33                              & 74.83                                  & 34.82                             & 77.53                             & 68.82                                   & 77.65                              \\
\rowcolor[HTML]{F5D7D6} 
FBQuant         & 3             & 128            & \textbf{69.11}                    & 47.87                              & 77.36                              & 77.76                                  & 49.32                             & 80.30                             & 71.98                                   & 79.17                              \\ \bottomrule
\end{tabular}
\end{table*}

\begin{table*}[h]
\caption{Zero-shot accuracy of Llama2-70B.}
\label{tab:ap3}
\begin{tabular}{@{}ccc|cccccccc@{}}
\toprule
\textbf{Method} & \textbf{Wbit} & \textbf{Group} & \multicolumn{1}{c}{\textbf{Avg.}} & \multicolumn{1}{c}{\textbf{Arc-c}} & \multicolumn{1}{c}{\textbf{Arc-e}} & \multicolumn{1}{c}{\textbf{HellaSwag}} & \multicolumn{1}{c}{\textbf{MMLU}} & \multicolumn{1}{c}{\textbf{PIQA}} & \multicolumn{1}{c}{\textbf{WinoGrande}} & \multicolumn{1}{c}{\textbf{BoolQ}} \\ \midrule
FP              & 16            & -              & \textbf{76.27}                    & 57.42                              & 82.74                              & 83.81                                  & 65.43                             & 82.70                             & 77.98                                   & 83.79                              \\ \midrule
RTN             & 4             & 128            & \textbf{75.87}                    & 56.31                              & 82.83                              & 83.00                                  & 64.36                             & 82.59                             & 78.30                                   & 83.70                              \\
GPTQ            & 4             & 128            & \textbf{75.85}                    & 56.80                              & 82.81                              & 82.68                                  & 64.98                             & 82.05                             & 78.10                                   & 83.56                              \\
AWQ             & 4             & 128            & \textbf{76.07}                    & 56.74                              & 82.66                              & 83.57                                  & 65.28                             & 82.75                             & 77.82                                   & 83.68                              \\
OmniQuant       & 4             & 128            & \textbf{76.10}                    & 57.01                              & 82.56                              & 83.88                                  & 65.21                             & 82.68                             & 77.80                                   & 83.56                              \\
CALDERA         & 4             & 128            & \textbf{76.01}                    & 57.77                              & 82.63                              & 83.64                                  & 64.94                             & 82.31                             & 76.82                                   & 83.95                              \\
SVDQuant        & 4             & 128            & \textbf{76.18}                    & 58.02                              & 82.41                              & 83.52                                  & 64.99                             & 82.86                             & 77.19                                   & 84.25                              \\
\rowcolor[HTML]{F5D7D6} 
FBQuant         & 4             & 128            & \textbf{76.19}                    & 57.32                              & 82.55                              & 83.79                                  & 65.38                             & 82.69                             & 77.82                                   & 83.74                              \\ \midrule
RTN             & 3             & 128            & \textbf{73.97}                    & 53.75                              & 80.85                              & 81.20                                  & 60.87                             & 81.88                             & 77.90                                   & 81.35                              \\
GPTQ            & 3             & 128            & \textbf{73.39}                    & 54.98                              & 79.62                              & 80.89                                  & 62.89                             & 79.74                             & 75.03                                   & 80.58                              \\
AWQ             & 3             & 128            & \textbf{74.82}                    & 55.38                              & 81.06                              & 82.54                                  & 63.54                             & 83.03                             & 76.16                                   & 82.04                              \\
OmniQuant       & 3             & 128            & \textbf{74.55}                    & 55.85                              & 80.88                              & 82.17                                  & 63.88                             & 81.00                             & 76.22                                   & 81.86                              \\
CALDERA         & 3             & 128            & \textbf{73.56}                    & 55.91                              & 79.97                              & 80.94                                  & 62.85                             & 79.66                             & 74.35                                   & 81.24                              \\
SVDQuant        & 3             & 128            & \textbf{74.61}                    & 57.00                              & 81.57                              & 81.98                                  & 62.19                             & 82.26                             & 75.22                                   & 82.05                              \\
\rowcolor[HTML]{F5D7D6} 
FBQuant         & 3             & 128            & \textbf{75.77}                    & 57.00                              & 82.10                              & 83.33                                  & 65.02                             & 82.24                             & 77.39                                   & 83.28                              \\ \bottomrule
\end{tabular}
\end{table*}

\begin{table*}[h]
\caption{Zero-shot accuracy of Llama3-8B.}
\label{tab:ap4}
\begin{tabular}{@{}ccc|cccccccc@{}}
\toprule
\textbf{Method} & \textbf{Wbit} & \textbf{Group} & \textbf{Avg.}  & \textbf{Arc-c} & \textbf{Arc-e} & \textbf{HellaSwag} & \textbf{MMLU} & \textbf{PIQA} & \textbf{WinoGrande} & \textbf{BoolQ} \\ \midrule
FP              & 16            & -              & \textbf{72.79} & 53.41          & 80.09          & 79.14              & 62.19         & 80.74         & 72.61               & 81.35          \\ \midrule
RTN             & 4             & 128            & \textbf{71.52} & 51.37          & 79.29          & 78.11              & 59.48         & 79.49         & 73.24               & 79.66          \\
GPTQ            & 4             & 128            & \textbf{66.12} & 49.18          & 61.95          & 70.61              & 58.16         & 75.51         & 72.35               & 75.10          \\
AWQ             & 4             & 128            & \textbf{71.78} & 52.54          & 79.00          & 78.86              & 61.60         & 79.08         & 72.43               & 78.96          \\
OmniQuant       & 4             & 128            & \textbf{71.81} & 52.80          & 78.91          & 79.15              & 61.53         & 79.02         & 72.41               & 78.85          \\
CALDERA         & 4             & 128            & \textbf{70.09} & 52.27          & 77.20          & 77.13              & 59.88         & 75.90         & 70.84               & 77.41          \\
SVDQuant        & 4             & 128            & \textbf{71.42} & 50.6           & 78.16          & 77.76              & 59.41         & 80.2          & 74.59               & 79.24          \\
\rowcolor[HTML]{F5D7D6} 
FBQuant         & 4             & 128            & \textbf{72.55} & 53.38          & 79.71          & 78.89              & 61.76         & 79.94         & 74.30               & 79.84          \\ \midrule
RTN             & 3             & 128            & \textbf{61.51} & 40.44          & 65.4           & 68.96              & 45.86         & 74.81         & 66.54               & 68.59          \\
GPTQ            & 3             & 128            & \textbf{63.63} & 40.96          & 67.68          & 72.54              & 49.12         & 73.72         & 68.27               & 73.12          \\
AWQ             & 3             & 128            & \textbf{67.83} & 47.61          & 75.84          & 74.48              & 49.42         & 78.89         & 70.88               & 77.68          \\
OmniQuant       & 3             & 128            & \textbf{67.97} & 48.92          & 75.24          & 74.72              & 49.64         & 78.91         & 70.72               & 77.63          \\
CALDERA         & 3             & 128            & \textbf{63.48} & 47.85          & 70.01          & 71.55              & 45.24         & 70.74         & 65.82               & 73.11          \\
SVDQuant        & 3             & 128            & \textbf{60.38} & 36.77          & 67.59          & 68.72              & 39.05         & 74.05         & 67.4                & 69.08          \\
\rowcolor[HTML]{F5D7D6} 
FBQuant         & 3             & 128            & \textbf{68.87} & 46.59          & 75.84          & 76.54              & 55.82         & 78.13         & 72.61               & 76.54          \\ \bottomrule
\end{tabular}
\end{table*}

\begin{table*}[h]
\caption{Zero-shot accuracy of Llama3-70B.}
\label{tab:ap5}
\begin{tabular}{@{}ccc|cccccccc@{}}
\toprule
\textbf{Method} & \textbf{Wbit} & \textbf{Group} & \multicolumn{1}{c}{\textbf{Avg.}} & \multicolumn{1}{c}{\textbf{Arc-c}} & \multicolumn{1}{c}{\textbf{Arc-e}} & \multicolumn{1}{c}{\textbf{HellaSwag}} & \multicolumn{1}{c}{\textbf{MMLU}} & \multicolumn{1}{c}{\textbf{PIQA}} & \multicolumn{1}{c}{\textbf{WinoGrande}} & \multicolumn{1}{c}{\textbf{BoolQ}} \\ \midrule
FP              & 16            & -              & \textbf{80.22}                    & 64.33                              & 86.99                              & 84.88                                  & 75.07                             & 84.49                             & 80.58                                   & 85.2                               \\ \midrule
RTN             & 4             & 128            & \textbf{78.67}                    & 60.32                              & 85.65                              & 84.78                                  & 72.58                             & 84                                & 78.85                                   & 84.53                              \\
GPTQ            & 4             & 128            & \textbf{79.51}                    & 63.71                              & 86.30                              & 84.11                                  & 74.35                             & 83.82                             & 79.92                                   & 84.42                              \\
AWQ             & 4             & 128            & \textbf{79.13}                    & 63.22                              & 86.22                              & 83.73                                  & 73.89                             & 83.64                             & 79.17                                   & 84.06                              \\
OmniQuant       & 4             & 128            & \textbf{79.37}                    & 63.82                              & 86.06                              & 84.37                                  & 74.14                             & 83.73                             & 79.33                                   & 84.15                              \\
CALDERA         & 4             & 128            & \textbf{79.24}                    & 63.44                              & 85.63                              & 84.09                                  & 73.58                             & 83.60                             & 79.87                                   & 84.46                              \\
SVDQuant        & 4             & 128            & \textbf{77.85}                    & 61.71                              & 84.55                              & 82.44                                  & 73.38                             & 82.49                             & 77.79                                   & 82.61                              \\
\rowcolor[HTML]{F5D7D6} 
FBQuant         & 4             & 128            & \textbf{79.74}                    & 63.94                              & 86.51                              & 84.36                                  & 74.57                             & 83.98                             & 80.15                                   & 84.71                              \\ \midrule
RTN             & 3             & 128            & \textbf{68.61}                    & 50.68                              & 78.79                              & 70.29                                  & 62.16                             & 80.25                             & 65.9                                    & 72.20                              \\
GPTQ            & 3             & 128            & \textbf{77.19}                    & 63.65                              & 85.12                              & 82.10                                  & 70.19                             & 80.28                             & 78.79                                   & 80.20                              \\
AWQ             & 3             & 128            & \textbf{77.69}                    & 62.34                              & 83.99                              & 81.59                                  & 73.49                             & 82.32                             & 76.69                                   & 83.47                              \\
OmniQuant       & 3             & 128            & \textbf{77.81}                    & 62.54                              & 82.73                              & 82.55                                  & 73.24                             & 81.87                             & 77.47                                   & 84.26                              \\
CALDERA         & 3             & 128            & \textbf{77.17}                    & 61.35                              & 81.32                              & 80.05                                  & 73.10                             & 82.57                             & 77.56                                   & 84.23                              \\
SVDQuant        & 3             & 128            & \textbf{78.47}                    & 63.10                              & 85.29                              & 82.40                                  & 74.08                             & 81.66                             & 79.14                                   & 83.64                              \\
\rowcolor[HTML]{F5D7D6} 
FBQuant         & 3             & 128            & \textbf{78.99}                    & 63.54                              & 85.31                              & 83.59                                  & 73.94                             & 83.34                             & 79.44                                   & 83.77                              \\ \bottomrule
\end{tabular}
\end{table*}

\begin{table*}[h]
\caption{Zero-shot accuracy of Qwen2.5-7B.}
\label{tab:ap6}
\begin{tabular}{@{}ccc|cccccccc@{}}
\toprule
\textbf{Method} & \textbf{Wbit} & \textbf{Group} & \textbf{Avg.}  & \textbf{Arc-c} & \textbf{Arc-e} & \textbf{HellaSwag} & \textbf{MMLU} & \textbf{PIQA} & \textbf{WinoGrande} & \textbf{BoolQ} \\ \midrule
FP              & 16            & -              & \textbf{74.25} & 51.02          & 80.43          & 78.94              & 71.94         & 79.71         & 73.01               & 84.68          \\ \midrule
RTN             & 4             & 128            & \textbf{71.55} & 48.14          & 77.05          & 75.84              & 71.26         & 77.26         & 70.51               & 80.79          \\
GPTQ            & 4             & 128            & \textbf{72.09} & 49.58          & 78.21          & 76.68              & 69.78         & 76.15         & 71.08               & 83.16          \\
AWQ             & 4             & 128            & \textbf{73.15} & 49.15          & 79.17          & 78.26              & 71.31         & 79.6          & 71.67               & 82.91          \\
OmniQuant       & 4             & 128            & \textbf{73.41} & 50.54          & 79.46          & 78.25              & 71.01         & 78.72         & 72.29               & 83.61          \\
CALDERA         & 4             & 128            & \textbf{68.78} & 45.54          & 73.50          & 75.61              & 64.13         & 72.83         & 70.03               & 79.83          \\
SVDQuant        & 4             & 128            & \textbf{73.85} & 51.88          & 80.22          & 78.26              & 71.07         & 80.47         & 72.14               & 82.91          \\
\rowcolor[HTML]{F5D7D6} 
FBQuant         & 4             & 128            & \textbf{67.21} & 47.01          & 71.55          & 69.12              & 65.3          & 74.81         & 65.04               & 77.61          \\ \midrule
RTN             & 3             & 128            & \textbf{70.73} & 50.34          & 78.32          & 74.6               & 65.3          & 78.13         & 66.93               & 81.47          \\
GPTQ            & 3             & 128            & \textbf{70.84} & 50.48          & 77.71          & 73.19              & 68.96         & 76.14         & 68.78               & 80.59          \\
AWQ             & 3             & 128            & \textbf{70.98} & 49.52          & 77.15          & 77.24              & 69.67         & 76.21         & 68.94               & 78.15          \\
OmniQuant       & 3             & 128            & \textbf{64.45} & 45.11          & 65.48          & 68.42              & 64.21         & 68.55         & 61.53               & 77.87          \\
CALDERA         & 3             & 128            & \textbf{70.79} & 49.49          & 77.15          & 73.95              & 67.16         & 77.09         & 69.69               & 81.01          \\
SVDQuant        & 3             & 128            & \textbf{72.16} & 51.02          & 78.54          & 76.87              & 69.31         & 79.54         & 71.74               & 78.13          \\
\rowcolor[HTML]{F5D7D6} 
FBQuant         & 3             & 128            & \textbf{78.99} & 63.54          & 85.31          & 83.59              & 73.94         & 83.34         & 79.44               & 83.77          \\ \bottomrule
\end{tabular}
\end{table*}
\end{document}